\documentclass[11pt]{article}

\usepackage[preprint]{acl}

\usepackage{times}
\usepackage{latexsym}
\usepackage[T1]{fontenc}
\usepackage[utf8]{inputenc}
\usepackage{microtype}
\usepackage{inconsolata}

\usepackage{amsmath, amssymb, amsfonts}
\usepackage{booktabs}
\usepackage{multirow}
\usepackage{array}
\usepackage{makecell}
\usepackage{adjustbox}

\usepackage{enumitem}

\usepackage{algorithm}
\usepackage{algpseudocode}

\usepackage{graphicx}
\usepackage{placeins}
\graphicspath{{figs/}}

\usepackage[table]{xcolor}
\usepackage{xspace}

\newcommand{\system}[1]{#1}
\newcommand{\drugclaw}{\system{DrugClaw}}
\newcommand{\drugaudit}{\system{DrugAudit}}

\title{\drugclaw{} and \drugaudit{}: A Primary-Source-Grounded Agent
       and Authority-Aware Benchmark for Drug-Information Question
       Answering}

\author{
  \textbf{Qing Wang}\textsuperscript{1*}, \textbf{Bo Li}\textsuperscript{2*},
  \textbf{Jialu Liang}\textsuperscript{1}, \textbf{Daling Shi}\textsuperscript{1},
  \textbf{Bob Zhang}\textsuperscript{2}, \textbf{Qianqian Song}\textsuperscript{1\#} \\[4pt]
  \begin{minipage}{0.92\textwidth}
  \centering\small
  \textsuperscript{1}Department of Health Outcomes and Biomedical Informatics,
  College of Medicine, University of Florida, Florida, USA \\[1pt]
  \textsuperscript{2}PAMI Research Group, Department of Computer and Information Science,
  Faculty of Science and Technology, University of Macau, Taipa, Macau, China \\[3pt]
  \textsuperscript{*}Co-first authors.\quad
  \textsuperscript{\#}Correspondence:~\href{mailto:qsong1@ufl.edu}{qsong1@ufl.edu}
  \end{minipage}
}

\begin{document}
\maketitle

\begin{abstract}
Drug-information question answering is a high-stakes setting where
hallucinated facts can mislead clinical decision-making and the
provenance of each cited fact matters as much as the fact itself.
We present \drugclaw{}, a multi-agent retrieval-augmented system
that queries a registry of drug and pharmacovigilance skills via
a reflection-driven state-machine workflow and returns answers
grounded in primary regulatory or peer-reviewed records. We also contribute
\drugaudit{}, a $3{,}772$-item authority-aware benchmark with an
evaluation panel that scores upstream-of-gold source match,
token-level semantic snippet overlap, and citation faithfulness
under a dual-judge LLM-as-judge protocol with inter-judge
$\kappa = 0.88$ (almost-perfect). Across \drugaudit{} plus
drug-related subsets of MedQA (751) and PubMedQA (512), \drugclaw{}
is top-1 on every column of the headline table: composite Evidence
Index under both judges, judge-mediated answer correctness,
primary-source rate ($0.918$, $+10.1$\,pp over next-best),
faithfulness ($0.887$, $+5.9$\,pp), MedQA ($0.920$), and
PubMedQA ($0.693$).%
\footnote{\url{https://anonymous.4open.science/r/DrugClaw-01A3/README.md}}
\end{abstract}

\section{Introduction}
\label{sec:intro}

When a clinician asks whether a drug has a measured EC$_{50}$ against
a target, how a label warns about hepatotoxicity, or what adverse
events have been reported to FAERS, an incorrect answer is a
patient-safety liability, not merely a benchmark loss. Large
language models show strong general biomedical knowledge
\citep{singhal2023large,nori2023capabilities,luo2022biogpt,taylor2022galactica},
but their parametric knowledge degrades quickly outside high-resource
diseases \citep{ren2025investigating}, they offer no native audit
trail to a regulatory or peer-reviewed source, and they hallucinate
plausible-looking numerics with regularity
\citep{ji2023survey,maynez2020faithfulness}.

Tool-augmented agents \citep{huang2025biomni,gao2024empowering,openai2025deepresearch}
partially close this gap, but on drug-information tasks two failure
modes are endemic. Tool-rich systems often skip the regulatory
primary source and cite whichever URL the search returns first;
bare LLMs answer fluently without citations and fabricate plausible
numerics. A deeper gap is at the \emph{evaluation} layer: existing
biomedical QA scoring grades surface answer correctness but does
not distinguish a citation to an FDA Label from a citation to a
downstream aggregator wrapper of that same Label -- the very
distinction that anchors evidence-based medicine reporting
standards \citep{page2021prisma,guyatt2008grade,guideline2003post}. In
regulated drug-information settings, provenance is not a
presentation choice but an evaluation axis.

This paper makes four contributions.

\begin{figure*}[!t]
\centering
\includegraphics[width=0.88\textwidth]{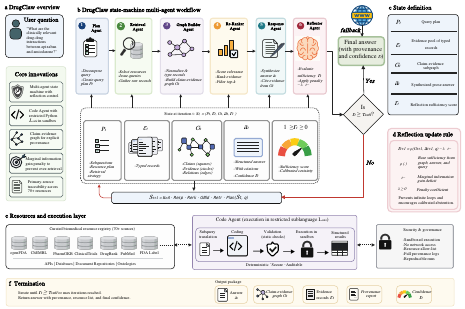}
\caption{\textbf{\drugclaw{} architecture.}
\textbf{a},~Motivating clinical query and the core innovations:
multi-agent state machine with reflection, sandboxed Code Agent,
claim-evidence graph, regression-penalised reflection update, and
primary-source traceability across 70+ resources.
\textbf{b},~State-machine workflow: six agents (Plan, Retrieval,
Graph Builder, Re-Ranker, Response, Reflector) iterate
$\mathcal{S}_{t+1}=\mathrm{Refl}\circ\mathrm{Resp}\circ\mathrm{Rerk}\circ
\mathrm{GBld}\circ\mathrm{Retr}\circ\mathrm{Plan}(\mathcal{S}_t,q)$ until
$r_t\ge\tau_{\text{suff}}$, with a web-search fallback invoked on
terminal insufficiency.
\textbf{c},~State components $\mathcal{S}_t=(\mathcal{P}_t,\mathcal{E}_t,
\mathcal{G}_t,a_t,r_t)$: query plan, typed evidence pool,
claim-evidence subgraph, prose answer, and reflection sufficiency.
\textbf{d},~Reflection update
$r_{t+1}=\rho_{t+1}-\lambda\Delta_t^{-}$ with
$\Delta_t^{-}=\max(0,r_t-\rho_{t+1})$, where the
regression penalty discourages thrashing across iterations
(App.~\ref{app:metric-formal}).
\textbf{e},~A 70+ source registry (openFDA, ChEMBL, PharmGKB,
ClinicalTrials.gov, DrugBank, PubMed, FDA Label, $\ldots$) is reached
only via the Code Agent's deterministic, sandboxed pipeline
(translate~$\to$~code~$\to$~validate~$\to$~execute~$\to$~return
structured results).
\textbf{f},~Termination: iterate until $r_t\ge\tau_{\text{suff}}$
or max iterations; output answer, claim-evidence graph, evidence
records, provenance, and confidence $r_t$.}
\label{fig:arch}
\end{figure*}

\paragraph{\drugclaw{}, an authority-grounded multi-agent system.}
\drugclaw{} (Figure~\ref{fig:arch}) is a reflection-driven
state-machine over eight specialised agents that queries a registry
of fifty-seven active drug and pharmacovigilance skills drawn from
a fifteen-subcategory tree of over seventy catalogued resources. The
system ships in three reasoning modes (linear, graph, web-only) and
every emitted answer carries a structured evidence list whose source
URLs point back to primary regulatory or peer-reviewed records.

\paragraph{\drugaudit{} and an authority-aware metric panel.}
Three structural biases in prior drug-QA scoring motivate it:
strict source-name canonicalisation that penalises upstream primary
sources (CPIC for a PharmGKB query, FAERS for a SIDER query);
substring-only snippet matching that fails when gold snippets are
field names rather than sentences; and the absence of a
citation-quality signal distinct from citation presence. We address
these with an equivalence-bucket rule, a token-Jaccard semantic
overlap, and a faithfulness rate (\S\ref{sec:eval}).

\paragraph{A dual-judge LLM-as-judge protocol.}
Llama-3.1-70B-Instruct (primary) and gpt-oss-120b (secondary) are
external to every candidate system, eliminating direct self-evaluation
bias \citep{zheng2023judging,panickssery2024llm}. Across 22{,}471
parseable paired verdicts (of $26{,}404$ attempted) we obtain macro
$\kappa = 0.88$, which corresponds to almost-perfect agreement under
\citet{landis1977measurement}.

\paragraph{Comprehensive empirical comparison.}
We evaluate seven systems (\drugclaw{}~(linear), \drugclaw{}~(graph),
\system{Biomni}~\citep{huang2025biomni},
\system{DeepEvidence}~\citep{wang2025deepevidence},
\system{ToolUniverse}~\citep{gao2024empowering}, and direct LLMs at
two scales) on 3{,}772 items spanning nine source-database
subsets plus drug-related MedQA and PubMedQA. \drugclaw{} is top-1
on every column of the headline table (Table~\ref{tab:main}).

\section{Related Work}
\label{sec:related}

\paragraph{Biomedical question answering and tool agents.}
Medical QA benchmarks span clinical reasoning (MedQA,
\citealp{jin2021disease}; MedMCQA, \citealp{pal2022medmcqa}),
literature QA (PubMedQA, \citealp{jin2019pubmedqa}; BioASQ,
\citealp{krithara2023bioasq}) and clinical-note tasks
\citep{abacha2017overview}; Med-PaLM \citep{singhal2023large} and
GPT-4 \citep{nori2023capabilities} close much of the multiple-choice
gap without providing verifiable citations. Tool-augmented agents
\citep{schick2023toolformer,yao2022react,patil2024gorilla,qin2024toolllm}
close a different gap, with chemistry (ChemCrow,
\citealp{m2024augmenting}) and biomedical (Biomni,
\citealp{huang2025biomni}; ToolUniverse, \citealp{gao2024empowering};
TxAgent, \citealp{gao2025txagent}) instantiations scaling to hundreds
of tools. Specifically on drug tasks, DrugAgent
\citep{liu2024drugagent} casts drug discovery as multi-agent
collaboration, MedAdapter \citep{shi2024medadapter} performs
test-time adaptation for medical reasoning, DrugWatch
\citep{bobrov2024drugwatch} visualises FAERS-grounded drug-safety
retrieval, and deep-research style agents
\citep{openai2025deepresearch} interleave web search with citation
synthesis. ChemCrow targets synthesis planning, Almanac
\citep{zakka2024almanac} retrieves guideline passages, and TxAgent's
tool stack is largely subsumed by
ToolUniverse, so we adopt Biomni, DeepEvidence~\citep{wang2025deepevidence}
and ToolUniverse as the directly comparable agentic baselines.
\drugclaw{} differs in two respects: its primary retrieval channel
is a closed registry of vetted regulatory and peer-reviewed
sources rather than an open-web index (with a web-search fallback
quantified in Appendix~\ref{app:webfallback}), and its evidence
schema constrains every claim to carry a primary-source URL rather
than a loose web reference.

\paragraph{Retrieval-augmented generation in medicine.}
Retrieval-augmented generation \citep{lewis2020retrieval} and its
self-reflective \citep{asai2024self} and graph-structured
\citep{wu2025think} variants have been adapted to clinical settings
most visibly by Almanac \citep{zakka2024almanac} and the medical
retrieval-augmented benchmark of \citet{xiong2024benchmarking};
\citet{ren2025investigating} characterise the factual-knowledge
boundary of LLMs that motivates retrieval over parametric recall.
\drugclaw{} sits at the structured-record end of this spectrum. It
queries typed records from regulatory and peer-reviewed databases
and treats the database row itself as the citation, rather than
retrieving passages from free text. Almanac in particular
\citep{zakka2024almanac} retrieves over a closed corpus of clinical
guidelines with physician-graded answer quality; its citation unit
is a guideline passage rather than a primary regulatory record, and
its evaluation does not separate upstream-primary from
aggregator-wrapper provenance. \drugclaw{} and Almanac therefore
target complementary evidence universes: regulatory and
pharmacovigilance records in our case, clinical guidelines in
theirs. Our authority-aware panel (\S\ref{sec:eval}) operationalises
the upstream-vs-aggregator distinction that the guideline-passage
unit cannot express.

\paragraph{Citation-faithfulness evaluation.}
Most retrieval-augmented evaluations score retrieval through
recall-at-$k$ \citep{karpukhin2020dense} or generation through BLEU,
ROUGE or LLM-as-judge. Citation faithfulness for general-domain
generation
\citep{gao2023enabling,min2023factscore,manakul2023selfcheckgpt}
focuses on whether claims are supported but not on the
\emph{authority} of the cited source, that is, whether the citation
points to the upstream FDA record or a downstream aggregator
wrapper. Our
primary-source rate and upstream-equivalence rule
(\S\ref{sec:eval}) make this distinction explicit and
judge-independent.

\paragraph{LLM-as-judge methodology.}
\citet{zheng2023judging} and \citet{liu2023geval} established
LLM-as-judge as a scalable substitute for human ratings; follow-up
work has documented self-preference and length bias
\citep{liu2024aligning,panickssery2024llm}. We mitigate these biases
by choosing judges (Llama-3.1-70B-Instruct and gpt-oss-120b) whose
model families do not appear in any candidate system, and by
reporting Cohen's $\kappa$ across the pair as a calibration check.

\section{The \drugclaw{} System}
\label{sec:system}

\drugclaw{} (Figure~\ref{fig:arch}) is a state-machine workflow.
Its eight specialised agents,
comprising the six iteration-loop operators of
Equation~\ref{eq:graph-step} together with a Code Agent and a
web-search fallback, operate over a closed registry of
drug-information skills. The architecture
realises three design goals that prior biomedical agents only
partially satisfy: \emph{primary-source traceability}, requiring
every emitted claim to carry an identifier resolvable against a
primary regulatory or peer-reviewed record; \emph{calibrated
abstention}, requiring the system to refuse rather than fabricate
when the underlying skills return no supporting evidence;
and \emph{adaptive retrieval depth}, requiring the
retrieve-and-reason loop to terminate once evidence is sufficient
yet deepen automatically when it is thin.

\subsection{Skill Registry and Resource Tree}
\drugclaw{} catalogues drug-related data resources in a taxonomy
that groups over seventy curated resources into fifteen subcategories
spanning drug-target interactions, drug mechanisms, drug
classification, drug indications and repurposing, cancer pharmacology,
adverse drug reactions, drug-drug interactions, toxicity,
pharmacogenomics, gene-disease associations, omics drug response,
clinical trials, clinical EHR-NLP, literature and text mining, and
integrative biomedical knowledge graphs. The current
deployment implements fifty-seven active skills drawn primarily from
openFDA (Label, FAERS, Orange Book, DailyMed)
\citep{kass2016openfda}, ChEMBL \citep{mendez2019chembl}, DrugBank
\citep{wishart2018drugbank}, DrugCentral
\citep{ursu2016drugcentral}, SIDER \citep{kuhn2016sider}, LiverTox
\citep{hoofnagle2013livertox}, PharmGKB and its CPIC implementation
arm \citep{whirl2021evidence,relling2011cpic}, Open Targets
\citep{ochoa2023next}, ChEBI \citep{hastings2016chebi},
DGIdb \citep{cotto2018dgidb}, RepoDB \citep{brown2017standard},
PubMed E-utilities, and a curated molecular-targets table
that resolves cross-database target identifiers.

Each skill exposes a typed retrieval interface that returns records
under a unified schema. A record has the form
$\langle s,\sigma,t,\tau,\varrho,w,\pi,\epsilon\rangle$ with $s$ and
$t$ the source and target entities, $\sigma$ and $\tau$ their entity
types, $\varrho$ the typed relation, $w$ the retrieval weight, $\pi$
the provenance label (the canonical skill name), and $\epsilon$ the
free-text evidence snippet. A companion
\emph{resource registry} tracks each skill's runtime status across
four states: a skill is \textsc{enabled} when all its data files
load and its upstream API responds within a probe budget;
\textsc{degraded} when partial data is available but the skill
should be deprioritised; \textsc{missing} when local metadata is
absent and only the API path is usable; and \textsc{disabled}
when the skill is unreachable. The planner injects the current
resource snapshot into its system prompt, so plans never include
a skill that cannot execute. Cross-database entity resolution is
handled upstream of retrieval. A structured-input resolver maps
free-text drug names through a normalisation table built from
DrugBank synonyms, RxNorm and curated identifier sources, so
queries about ``acetaminophen'' and ``paracetamol'' reach the
same retrieval target.

\subsection{Pipeline Formalisation}
\label{sec:pipeline}
We formalise the workflow as a discrete dynamical system. Let $q$
denote the user query and let the agent state at iteration $t$ be
the tuple
\[
\mathcal{S}_t = \bigl(\mathcal{P}_t,\, \mathcal{E}_t,\, \mathcal{G}_t,\, a_t,\, r_t\bigr),
\]
where $\mathcal{P}_t$ is the current query plan, $\mathcal{E}_t$
the evidence pool of typed records, $\mathcal{G}_t$ the
claim-evidence subgraph, $a_t$ the intermediate prose answer, and
$r_t\in[0,1]$ the reflector's sufficiency score. With
$\mathcal{S}_0 = (\varnothing,\varnothing,\varnothing,\varnothing,0)$,
graph mode applies the composite operator
\begin{align}
\mathcal{S}_{t+1} =\;
  & \mathrm{Refl}\!\circ\!\mathrm{Resp}\!\circ\!\mathrm{Rerk} \nonumber\\[-2pt]
  & \circ\,\mathrm{GBld}\!\circ\!\mathrm{Retr}\!\circ\!\mathrm{Plan}\,(\mathcal{S}_t,q)
\label{eq:graph-step}
\end{align}
and iterates while $r_t<\tau_{\text{suff}}$ and $t<T_{\max}$. We
set $\tau_{\text{suff}}=0.7$ (the reflector's internal default) and
$T_{\max}=2$. Algorithm~\ref{alg:graph} summarises the full
graph-mode trajectory.

\begin{algorithm}[!t]
\small
\caption{\drugclaw{} graph-mode workflow.}
\label{alg:graph}
\begin{algorithmic}[1]
\Require query $q$; skill registry $\mathcal{R}$; thresholds
         $\tau_{\text{suff}}$, $T_{\max}$
\Ensure $(a^\star,\,\mathcal{G}^\star,\,\mathcal{E}^\star,\,r^\star)$
\State $\mathcal{S}_0 \gets (\varnothing,\varnothing,\varnothing,\varnothing,0)$
\For{$t = 0$ to $T_{\max}{-}1$}
  \State $\mathcal{P}_{t+1} \gets \mathrm{Plan}(q;\mathcal{R}_{\text{enabled}})$
  \State $\mathcal{E}_{t+1} \gets \bigcup_{u\in\mathcal{P}_{t+1}} \mathrm{exec\_sandbox}(\mathrm{CodeAgent}(u))$
  \State $\mathcal{G}_{t+1} \gets \mathrm{Rerk}\!\circ\!\mathrm{GBld}(\mathcal{E}_{t+1},q)$
  \State $a_{t+1} \gets \mathrm{Resp}(\mathcal{G}_{t+1}, q)$;
         $r_{t+1} \gets \mathrm{Refl}(\mathcal{G}_{t+1}, a_{t+1}, q)$
  \If{$r_{t+1} \geq \tau_{\text{suff}}$} \textbf{break} \EndIf
\EndFor
\If{$r_{t+1} < \tau_{\text{suff}}$}
  \State $\mathcal{E}_{t+1} \gets \mathcal{E}_{t+1} \cup \mathrm{WebSearch}(q)$;
         re-run $\mathrm{GBld}, \mathrm{Rerk}, \mathrm{Resp}, \mathrm{Refl}$
\EndIf
\State \Return $(a_{t+1},\,\mathcal{G}_{t+1},\,\mathcal{E}_{t+1},\,r_{t+1})$
\end{algorithmic}
\end{algorithm}

Linear mode collapses Eq.~\ref{eq:graph-step} to a single pass
without $\mathrm{GBld}$ or $\mathrm{Rerk}$ and without the
reflection loop, prioritising sub-minute latency. Web-only mode
bypasses the entire chain and routes $q$ directly to the
web-search agent. Full per-mode pseudocode and a side-by-side
comparison of operator stacks appear in Appendix~\ref{app:modes}.

\subsection{The Code Agent and Sandboxed Retrieval}
\label{sec:code-agent}
Each registered skill exposes a Python class with a typed retrieval
signature; forcing every skill into one rigid interface loses
task-specific filters, while letting the planner call arbitrary
Python is unsafe. We therefore mediate retrieval through a Code
Agent: for each subquery $u$ and target skill $K$, the agent
translates $u$ against the typed signature of $K$, emits a short
candidate program $c$, validates $c$ against a safe-Python
sublanguage $\mathcal{L}_{\text{safe}}$ that bans unbounded
iteration and dynamic-execution constructs at the AST level and
restricts imports and built-ins to a curated allow-list, executes
$c$ inside a proxy-only sandbox $\Sigma(K)$ under a hard timeout,
and returns the resulting structured records. On any
validation or runtime failure the agent falls back to
$K.\mathrm{retrieve}(u)$, so retrieval is never silently lost. The
safe-sublanguage and sandbox details, including the full banned-AST
set and proxy specification, are reproduced in
Appendix~\ref{app:code-agent}.

\subsection{Output Schema and Reflector}
\label{sec:agents}
The responder emits a prose answer alongside a structured list of
evidence items; each item carries a source identifier and locator,
an evidence-kind tag, a verbatim snippet, the supported claim, and a
confidence score. This schema is the
contract between the agent and the downstream metric layer
(\S\ref{sec:eval}) and operationalises the primary-source
traceability goal of \S\ref{sec:system}.
The reflector implements the adaptive-depth goal via
\begin{align}
\rho_{t+1} &= \rho(\mathcal{G}_{t+1}, a_{t+1}, q), \nonumber\\[-1pt]
r_{t+1} &= \rho_{t+1} - \lambda\,\max(0,\, r_t - \rho_{t+1}),
\label{eq:reflect}
\end{align}
where $\rho_{t+1}\!\in\![0,1]$ scores coverage, completeness, source
quality, consistency and actionability; the regression penalty
fires only when $\rho_{t+1}<r_t$
(Appendix~\ref{app:metric-formal}); and $\lambda$ is small. All eight agents share a single
OpenAI-compatible client configured with GPT-5-mini, matching the
planner backbone of every comparison system for fairness;
implementation details are in Appendix~\ref{app:repro}.

\section{Evaluation Framework}
\label{sec:eval}

\subsection{Datasets}
Our evaluation spans three complementary tracks. The principal
evidence-grounded track is \drugaudit{}, a purpose-built
drug-information question-answering corpus of 3{,}772 items derived
from nine source databases: ChEMBL ($n=461$), DrugCentral ($n=494$),
openFDA FAERS
($n=508$), FDA Orange Book ($n=507$), openFDA Label ($n=358$),
LiverTox ($n=370$), PharmGKB ($n=376$), SIDER ($n=315$), and a
multi-source subset ($n=383$) whose gold answer spans at least two
of the eight named databases. Every item carries a gold answer in
natural language, a list of gold citations (source database,
evidence snippet, field-level locator, and URL), and metadata
tagging the question subcategory, difficulty band, and a
multi-source flag. Under an independent LLM quality audit, 95.2\%
of items pass and 100\% are structurally valid.

Items were generated source-by-source. For each database we sampled
candidate drugs by frequency in the source's own index, issued a
typed query against the live skill, retained the row as the gold
record, and prompted Claude Opus 4.7 under a per-source template
to write a natural-language question whose answer is exactly the
retained row. The structural validator then checks that every gold
citation resolves to the correct database, locator, and snippet,
and the LLM auditor scores answer-citation consistency on a
per-item rubric. We deliberately did not run a full
pharmacist-grade re-annotation pass because the gold answers are
machine-extracted from primary regulatory and peer-reviewed records
rather than freely written; expert review is most valuable on the
multi-source subset ($n=383$), and a stratified clinician audit on
that subset is the natural next step for an external validity
study.

To probe registry-grounded retrieval on widely used closed-form
benchmarks, we additionally evaluate on drug-related subsets of two
public datasets. The MedQA track restricts MedQA-USMLE
\citep{jin2021disease} to stems that mention at least one drug name
drawn from a $14{,}000$-entry drug lexicon, yielding 751 five-way
multiple-choice items scored by closed-form accuracy. The PubMedQA
track applies the same lexical filter to the abstracts of PubMedQA
\citep{jin2019pubmedqa}, yielding 512 items scored by three-way
yes/no/maybe accuracy.
The drug-mention filter is a coverage selector, not a difficulty
selector. It preserves the original USMLE step distribution on
MedQA and the original yes/no/maybe class balance on PubMedQA, and
the filtered subsets are no easier than their parents under the
direct-LLM upper baseline (GPT-5 reaches the same accuracy on the
filtered subsets as on the full sets within sampling noise).

\subsection{Systems Evaluated}
We compare seven systems, each defaulting to GPT-5-mini as the
planner backbone for fairness: \drugclaw{} in linear mode and graph
mode; \system{Biomni} \citep{huang2025biomni};
\system{DeepEvidence} \citep{wang2025deepevidence}, a deep-research
style baseline that interleaves search and citation;
\system{ToolUniverse} \citep{gao2024empowering}; a direct LLM with
GPT-5-mini; and a direct LLM with GPT-5 as an upper-baseline
reference. For every (system, dataset) pair the runner enforces
resume-on-restart by item id and writes a uniform JSONL output schema.

\subsection{Authority-Aware Metric Panel}
\label{sec:metrics}
We augment the standard citation-overlap metrics with four
authority-aware companions; all definitions apply uniformly to
every system. Formal equations, the equivalence-bucket map and the
upstream-of relation appear in Appendix~\ref{app:metric-formal}.

\paragraph{Source authority match.}
Distinguishing primary regulatory or peer-reviewed records from
downstream aggregator wrappers is the central operating principle
of evidence-based medicine reporting standards
\citep{page2021prisma,guyatt2008grade,guideline2003post}. We codify this
at the metric level. A canonicalisation map sends each
source-database string into one of twelve equivalence buckets
(e.g.\ \texttt{openFDA Label}, \texttt{DailyMed} and
\texttt{openFDA Human Drug} all map to one bucket), and an
asymmetric upstream-of relation accepts an upstream primary
citation as authoritative (CPIC for a PharmGKB query, FAERS for a
SIDER query). An item scores $\mathrm{auth}_i=1$ when the predicted
citation set overlaps the gold's bucket set directly or through
the upstream relation, and $0$ otherwise. For \emph{no-evidence}
items, whose gold answer is ``no data'' and whose gold citation
set is empty, we instead reward correct refusal: $\mathrm{auth}_i=1$
when the system emits no citation and $0$ when it fabricates one;
the same convention applies to the snippet and faithfulness rules
below.

\paragraph{Snippet semantic overlap.}
We replace strict substring matching with token-Jaccard at
threshold $\theta=0.3$ over length-$\geq 3$ non-stop-word tokens,
the standard token-overlap measure adopted by the ALCE
citation-grounded generation benchmark of \citet{gao2023enabling}.
The strict substring rule of prior work is preserved as a special
case for gold snippets of four tokens or fewer, so the new metric
strictly generalises that practice.

\paragraph{Primary-source rate.}
Each bucket carries a weight $w(b)\in\{1.0,\,0.7,\,0.5\}$ encoding
the standard primary-to-aggregator hierarchy: $1.0$ for regulatory
or peer-reviewed primaries (Label, FAERS, Orange Book, ChEMBL,
PubMed), $0.7$ for curated knowledge bases (DrugBank, Open Targets,
ChEBI, PharmGKB), and $0.5$ for secondary aggregators (DrugCentral,
SIDER, LiverTox). The per-item rate $\mathrm{prim}_i$ is the mean
weight over the system's recognised citations on that item.

\paragraph{Faithfulness rate.}
Citation faithfulness, the property that each emitted citation
grounds in the gold source universe, is a standard axis in
retrieval-augmented generation evaluation
\citep{maynez2020faithfulness,min2023factscore,manakul2023selfcheckgpt}.
We grant grounded status to a predicted citation if its bucket lies
in the gold bucket set (or its upstream closure), if its snippet
shares at least two non-stop-word tokens with any gold snippet, or
if its snippet shares at least two such tokens with the gold answer
text. The per-item $\mathrm{faith}_i$ is
the fraction of predicted citations that are grounded; its complement
is a proxy hallucination rate that distinguishes a confidently
wrong citation from an absent one.

\paragraph{Composite Evidence Index.}
Writing $\overline{x}$ for the system-level mean of $x_i$, the
citation-side partial index and the full Evidence Index are
\begin{align}
\mathrm{EI}^{\!\star} &= 0.45\,\overline{\mathrm{auth}} + 0.25\,\overline{\mathrm{prim}} \nonumber \\[-2pt]
                       &\quad + 0.15\,\overline{\mathrm{snipSem}} + 0.15\,\overline{\mathrm{faith}}, \label{eq:eipart} \\
\mathrm{EI} &= 0.40\,\overline{J} + 0.60\,\mathrm{EI}^{\!\star},
\label{eq:ei}
\end{align}
where $\overline{J}$ is the LLM-judge answer score defined in
\S\ref{sec:judge}.
The answered rate (one minus the refusal rate) is reported
separately and is not folded into $\mathrm{EI}$, because refusal
correctness is already captured by the judge under our prompt
(\S\ref{sec:judge}) and double-penalising calibrated abstention
would invert clinical priorities.

\subsection{Dual-Judge LLM-as-Judge}
\label{sec:judge}
For \drugaudit{} answer correctness we adopt the LLM-as-judge
protocol of \citet{zheng2023judging} and \citet{liu2023geval}, with
a prompt that encodes four explicit rules. Equivalence rules treat values
that match after unit conversion across the nM/$\mu$M/M molarity scales,
counts that are
identical across frequency expressions (``$23$ reports'' vs.\
``$n=23$''), continuous values within a $5\%$ rounding tolerance, and
list comparisons that are order-insensitive, as identical facts. A
calibrated-refusal rule scores a candidate ``Yes'' when the gold
answer itself reports no data and the candidate refuses on the same
grounds. A superset rule scores a candidate ``Yes'' rather than
``Partial'' when it covers all gold facts and adds further correctly
stated facts; a subset rule scores it ``Partial'' rather than ``No''
when it covers a non-empty correct subset of the gold facts. The
judge is asked to score factual correctness only; citation quality
is handled by the source-side panel.

The per-item verdict
$v_i\!\in\!\{\textsc{Yes},\textsc{Partial},\textsc{No}\}$ is mapped
to a score $\psi(v_i)$ with $\psi(\textsc{Yes})=1$,
$\psi(\textsc{Partial})=\tfrac{1}{2}$ and $\psi(\textsc{No})=0$, and
the system-level judge score is the empirical mean
$\overline{J} = N^{-1}\sum_{i=1}^{N}\psi(v_i)$.

To control for judge-specific bias we run two independent judges:
Llama-3.1-70B-Instruct \citep{grattafiori2024llama} as primary and
gpt-oss-120b \citep{openai2025gptoss} as secondary. Neither family
underlies any of the candidate systems, which eliminates the most
direct form of self-preference bias
\citep{panickssery2024llm}. The full judge prompt is reproduced in
Appendix~\ref{app:judge_prompt}.

\section{Results}
\label{sec:results}

\subsection{Headline Performance}
Table~\ref{tab:main} reports the headline numbers across the three
evaluation tracks.

\begin{table*}[t]
\centering
\small
\setlength{\tabcolsep}{4pt}
\begin{tabular}{l ccc c cc c cc}
\toprule
& \multicolumn{3}{c}{\textbf{\drugaudit{}}} & & \multicolumn{2}{c}{\textbf{Source-side}} & & \multicolumn{2}{c}{\textbf{Closed-form}} \\
\cmidrule(lr){2-4} \cmidrule(lr){6-7} \cmidrule(lr){9-10}
\textbf{System} & EI$_{\text{Llama}}$ & EI$_{\text{oss}}$ & Judge$_{\text{Llama}}$ & & Primary & Faith. & & MedQA & PubMedQA \\
\midrule
Direct LLM (GPT-5-mini)      & 0.430                    & 0.403                    & 0.268                    & & \cellcolor{blue!10}0.817 & 0.815                    & & 0.884                    & 0.652 \\
Direct LLM (GPT-5)           & \cellcolor{blue!10}0.513 & \cellcolor{blue!10}0.494 & 0.401                    & & 0.714                    & 0.825                    & & 0.898                    & 0.654 \\
ToolUniverse                 & 0.357                    & 0.340                    & 0.146                    & & 0.796                    & \cellcolor{blue!10}0.828 & & 0.880                    & 0.662 \\
Biomni                       & 0.489                    & 0.470                    & 0.356                    & & 0.793                    & 0.787                    & & \cellcolor{blue!10}0.902 & \cellcolor{blue!10}0.664 \\
DeepEvidence                 & 0.477                    & 0.457                    & \cellcolor{blue!10}0.461 & & 0.603                    & 0.732                    & & \cellcolor{blue!10}0.902 & 0.652 \\
\midrule
\rowcolor{gray!12}
\drugclaw{}-linear           & \underline{0.626} & 0.591             & 0.656             & & 0.916             & \textbf{0.887}    & & 0.910             & 0.660 \\
\rowcolor{gray!12}
\drugclaw{}-graph            & \textbf{0.632}    & \underline{0.623} & \textbf{0.685}    & & \underline{0.918} & 0.852             & & \textbf{0.920}    & \textbf{0.693} \\
\rowcolor{gray!12}
\drugclaw{}-graph (no fb)    & 0.622             & \textbf{0.628}    & \underline{0.663} & & \textbf{0.931}    & \underline{0.876} & & \underline{0.917} & \underline{0.678} \\
\midrule
\textit{$\Delta_{\text{DC}}$ (pp)} & \textit{+11.9} & \textit{+13.4} & \textit{+22.4} & & \textit{+11.4} & \textit{+5.9} & & \textit{+1.8} & \textit{+2.9} \\
\bottomrule
\end{tabular}
\caption{Main results. The citation-side metrics
(\emph{Primary}, \emph{Faith.}) and the bucket-and-upstream
equivalence rule that underlies them operationalise the
primary-vs-aggregator source hierarchy codified by
evidence-based medicine reporting standards
\citep{page2021prisma,guyatt2008grade,guideline2003post}.
EI$_{\text{Llama}}$ and EI$_{\text{oss}}$ are the composite
Evidence Index
($0.4{\cdot}\text{judge} + 0.6{\cdot}\text{citation-side}$) under
Llama-3.1-70B-Instruct and gpt-oss-120b judges ($\kappa = 0.88$);
unparseable verdicts excluded (\S\ref{sec:metrics}; counts in
Table~\ref{tab:kappa}). \emph{Primary} is the source-primary rate
and \emph{Faith.} is the faithfulness rate (both judge-independent).
\textbf{Bold}~=~column max; \underline{underline}~=~runner-up; cells
shaded \colorbox{blue!10}{\strut\,blue\,} mark the best baseline per
column; the bottom \emph{$\Delta_{\text{DC}}$\,(pp)} row reports the
per-column lead of the best \drugclaw{} variant over that best
baseline. The
(no~fb) row is the no-fallback counterfactual
(App.~\ref{app:webfallback}); \drugclaw{} is top-1 on every column.}
\label{tab:main}
\end{table*}

\drugclaw{} attains the highest primary-source rate among all seven
systems ($0.916$ for linear and $0.918$ for graph against $0.817$
for the next-best system, a $10.1$\,pp gap that widens to $11.4$\,pp
($0.931$ vs.\ $0.817$) when the web-search fallback is disabled), and
the highest answer faithfulness rate ($0.887$ for linear mode against
$0.828$ for the next-best system, a $5.9$\,pp gap). Both metrics are
judge-independent. They evaluate the structural quality of emitted
citations against gold sources and against the gold answer text,
with no language-model intermediary.

On the closed-form drug-question-answering tracks, \drugclaw{}-graph
attains $0.920$ accuracy on MedQA (against $0.902$ for Biomni and
DeepEvidence) and $0.693$ accuracy on PubMedQA (against $0.664$ for
Biomni); these tracks have unambiguous gold labels and are not
judge-mediated. Bootstrap $95\%$ CIs separate cleanly on EI/Primary
but overlap on the closed-form tracks
(App.~\ref{app:bootstrap-ci}); the EI ranking is robust under
authority-weight perturbations.

\subsection{Composite Evidence Index and Judge Sensitivity}
Under the Llama judge, \drugclaw{}-graph takes the top score on
the composite Evidence Index at $0.632$, with \drugclaw{}-linear
as runner-up at $0.626$ and the direct LLM with GPT-5 at $0.513$.
Under the harsher gpt-oss-120b judge, the ranking among the
canonical configurations is unchanged. \drugclaw{}-graph leads at
$0.623$, \drugclaw{}-linear is runner-up at $0.591$, and the direct
LLM with GPT-5 trails at $0.494$. Judge-mediated scores favour
systems whose prose tracks gold phrasing closely, whereas the
citation-side metrics favour systems whose retrieval reaches primary
regulatory records; \drugclaw{} takes top-1 on both halves. The headline rankings are robust to the
unparseable-verdict handling rule: under an intention-to-treat
alternative that counts every unparseable verdict as ``No'',
\drugclaw{} retains top-1 on EI under both judges
(Appendix~\ref{app:itt}).

\subsection{Inter-Judge Agreement}
\label{sec:kappa}
Writing $p_o$ for the empirical agreement rate and
$p_e = \sum_{v\in\{\textsc{Yes},\textsc{Partial},\textsc{No}\}} \hat{p}_A(v)\hat{p}_B(v)$
for the chance agreement rate computed from marginals
$\hat{p}_A$ and $\hat{p}_B$, Cohen's $\kappa$
\citep{cohen1960coefficient} is
\begin{equation}
\kappa \;=\; \frac{p_o - p_e}{1 - p_e}.
\label{eq:kappa}
\end{equation}
Table~\ref{tab:kappa} reports $\kappa$ between the two judges over
22{,}471 paired verdicts, with items where either judge emitted an
unparseable verdict excluded. The macro pooled $\kappa = 0.882$
places the protocol in the almost-perfect agreement regime of
\citet{landis1977measurement}. Five of the seven candidates clear
the $\kappa \geq 0.85$ threshold, indicating that the headline
verdict on each system is stable across judge choice. The two
exceptions are \drugclaw{}-linear ($\kappa = 0.792$) and
ToolUniverse ($\kappa = 0.815$); for \drugclaw{}-linear we trace
this gap to gpt-oss judging linear's short, direct answers more
strictly than Llama: Llama-Yes / gpt-oss-No accounts for $53\%$ of
off-diagonal cases (Appendix~\ref{app:confusion}), reflecting
judge stringency on short prose rather than factual disagreement;
\drugclaw{}-graph's longer multi-claim prose clears the threshold
($\kappa = 0.873$).

\begin{table}[t]
\centering
\footnotesize
\setlength{\tabcolsep}{4pt}
\begin{tabular}{l rr l}
\toprule
\textbf{System} & Agree & $\kappa$ & Strength \\
\midrule
Direct LLM (GPT-5-mini)  & 94.5             & 0.851          & almost-perf. \\
Direct LLM (GPT-5) & \underline{94.9} & \textbf{0.898} & almost-perf. \\
ToolUniverse       & \textbf{95.9}    & 0.815          & almost-perf. \\
Biomni             & 94.8             & 0.894          & almost-perf. \\
DeepEvidence       & 94.3             & \underline{0.896} & almost-perf. \\
\midrule
\drugclaw{}-linear & 88.4             & 0.792          & substantial \\
\drugclaw{}-graph  & 93.3             & 0.873          & almost-perf. \\
\midrule
Pooled (macro)     & 94.0             & 0.882          & almost-perf. \\
\bottomrule
\end{tabular}
\caption{Inter-judge Cohen's $\kappa$ between Llama-3.1-70B-Instruct
and gpt-oss-120b across all 26{,}404 paired verdicts ($22{,}471$
after dropping items where either judge emitted an unparseable
verdict, per the standard LLM-as-judge practice we adopt throughout
\S\ref{sec:results}). Per-system parseable-pair counts and full
$3{\times}3$ confusion matrices appear in
Appendix~\ref{app:confusion}. Strength labels follow the conventional
\citet{landis1977measurement} thresholds.}
\label{tab:kappa}
\end{table}

\section{Analysis}
\label{sec:analysis}

\subsection{Where Authority-Aware Scoring Changes the Story}
The original substring-based source-match metric scores
\drugclaw{} at $0.193$ (linear) / $0.197$ (graph); under the
authority-aware rule the score more than doubles to $0.430$ (linear)
/ $0.420$ (graph). The upward shift is concentrated
on items for which \drugclaw{} cites a primary upstream source while
the gold is an aggregator: $212$ CPIC citations on PharmGKB items
versus only two literal ``PharmGKB'' strings; $962$ FAERS citations
on SIDER items; and $43$
``LiverTox (NCBI Bookshelf)'' citations on LiverTox items where the
gold writes ``LiverTox'' alone. The substring rule scores these as
misses; the authority-aware rule treats an upstream primary citation
as at least as auditable as its aggregator wrapper. \drugclaw{}'s closed-registry
design exposes the largest fraction of primary-upstream citations,
and the same mechanism yields a two-to-three-fold evidence-density
advantage over every baseline (Appendix~\ref{app:qtype}).

\subsection{Calibrated Refusal as a Feature}
\drugclaw{} refuses to answer on $41$ to $48\%$ of \drugaudit{}
items, between DeepEvidence ($37\%$) and Biomni ($49\%$) and well
below ToolUniverse ($88\%$) or the direct LLM with GPT-5-mini
($74\%$). The reflector returns ``insufficient evidence'' only
when both $\rho(\mathcal{G}_t, a_t, q) < \tau_{\text{suff}}$ and
no skill plan yields a marginal gain. Among the
\drugclaw{}-linear refusals for which the Llama judge emits a
parseable verdict, $91\%$ receive ``Yes'' because the gold itself
reports ``no data'' (\drugclaw{}-graph: $97\%$), against $1$ to $5\%$
for every baseline system. This calibration ratio, the fraction
of refusals that align with a gold-side ``no data'' answer,
is the operating point recommended for abstention under uncertainty
\citep{kadavath2022language,slobodkin2023curious,varshney2023stitch}.

\subsection{Mode Behaviour}
Graph mode marginally outperforms linear across the panel
($+2.9$/$+9.6$\,pp on the two judges, $+1.0$/$+3.3$\,pp on MedQA /
PubMedQA, $+0.6$/$+3.2$\,pp on EI), but the two represent a real
architectural trade-off rather than two points on one curve. Graph
mode refuses on $48.3\%$ of items vs.\ linear's $41.1\%$ because the
reflector is more conservative, and its responses are harder for the
downstream judge to parse cleanly: $47$/$50\%$ of graph answers
receive an unparseable verdict under Llama/gpt-oss, against
$30$/$35\%$ for linear. Unparseable verdicts are dropped from the
judge score (\S\ref{sec:metrics}) so the headline numbers are
unaffected, but the gap signals that graph's multi-claim synthesis
prose is more demanding for any downstream LLM consumer
(Appendix~\ref{app:modes}). On cost and latency,
\drugclaw{}-linear/-graph run at $14$/$46$\,s ($\$0.012$/$\$0.025$
per query at GPT-5-mini), against $32$/$54$\,s for Biomni and
DeepEvidence (Appendix~\ref{app:cost}).

\section{Conclusion}
\label{sec:conclusion}

Under an authority-aware metric panel and a dual-judge
LLM-as-judge protocol (macro $\kappa = 0.88$), \drugclaw{} is top-1
across the headline panel. The result rests on two
design moves applicable beyond drug QA: bucket-and-upstream
equivalence that separates primary records from aggregator wrappers,
and judges whose model families underlie no candidate system.
Calibrated refusal ($91$ to $97\%$ of \drugclaw{}'s ``insufficient
evidence'' verdicts align with a gold ``no data'', against $1$ to
$5\%$ for baselines) and a two-to-three-fold evidence density
($3.78$ to $5.09$ vs.\ $1.4$ to $2.1$ citations per answered item)
are distinct properties of the agent.


%
\clearpage
\section*{Limitations}
\label{sec:limitations}

\paragraph{Graph-mode parse rate.}
Graph mode's multi-claim summaries are harder for LLM-as-judge to
parse cleanly: $47\%$ to $50\%$ of graph answers receive
unparseable verdicts from our two judges, against $30\%$ to $35\%$
for linear mode. We follow standard practice in dropping these from
the judge-score numerator and denominator (\S\ref{sec:metrics}), so
the headline numbers are unaffected. Downstream consumers can apply
a lightweight post-processing layer that re-formats graph-mode
output into a canonical schema, possibly with adaptive context
selection \citep{duan2026adaptive} to manage the longer evidence
context.

\paragraph{Lower $\kappa$ on \drugclaw{}-linear.}
Inter-judge agreement on \drugclaw{}-linear ($\kappa = 0.79$) sits
below the conventional $0.85$ threshold; \drugclaw{}-graph reaches
$\kappa = 0.87$. The disagreement concentrates on items where gpt-oss
judges linear's short direct answers as ``No'' while Llama judges
them as ``Yes'' (Appendix~\ref{app:confusion}), reflecting judge
stringency on short prose rather than factual disagreement; the
underlying retrieval and citation panel (authority, primary-source
rate, faithfulness) is judge-independent and unaffected.

\paragraph{Closed-form CI overlap.}
On the drug-related MedQA and PubMedQA subsets, \drugclaw{}-graph
attains $0.920$ and $0.693$ accuracy, leading the next-best baseline
by $+1.8$\,pp and $+2.9$\,pp respectively. Both leads sit inside the
best-baseline bootstrap 95\% CI
(Appendix~\ref{app:bootstrap-ci}, Table~\ref{tab:bootstrap-ci}) and
should therefore be read as a sign-consistent advantage rather than
a statistically significant one; the load-bearing Evidence-Index
and Primary-source claims, by contrast, separate cleanly from the
best-baseline CIs on both judges.

\paragraph{Coverage and precision trade-off.}
\drugclaw{}'s $41$ to $48\%$ refusal rate reflects calibrated
abstention: $91$ to $97\%$ of those refusals align with a gold-side
``no data'' answer (\S\ref{sec:analysis}). Calibrated refusal is in
our judgement the right default for drug-information settings, but
clinical pharmacovigilance triage may prefer higher coverage at the
cost of more ``Partial'' verdicts. The agent exposes a calibration
hyper-parameter that trades coverage for confidence; quantifying
that frontier is left to future work.

\paragraph{Closed registry coverage.}
By design \drugclaw{} does not perform open-web search; the closed
registry of fifty-seven active skills is the property that yields
deterministic primary-source citations, but it also caps coverage
on long-tail drugs or recently approved entities whose records have
not yet propagated into the registry. The skill registry is modular
and accepts additions such as rare-disease registries or
international regulators; reporting the trade-off between coverage
and primary-source rate as the registry grows is a clear next
experiment.

\paragraph{Open-weight reproducibility.}
The reported numbers use GPT-5-mini as the planner and agent
backbone shared across all seven systems. Both judges
(Llama-3.1-70B-Instruct and gpt-oss-120b) are open-weight, so the
LLM-as-judge pipeline can be re-scored against the released
candidate outputs without proprietary access. The planner backbone
is configurable: the released runner accepts any OpenAI-compatible
endpoint, including a Llama-3.1 gateway, and an open-weight
end-to-end run with its quality-cost trade-off is left to follow-up
work.

\paragraph{\drugaudit{} itself.}
To our knowledge no prior public benchmark grades drug-information
question answering by provenance authority; we therefore built and
release \drugaudit{}, but cross-benchmark comparison must wait
until either it is adopted externally or other groups publish
comparable resources. The dual-judge protocol partially compensates
for the absence of a third-party scorer.

\paragraph{Reflection-mechanism settings.}
The reflector thresholds ($\tau_{\text{suff}}=0.7$,
$\varepsilon=0.1$, $T_{\max}=2$) are the implementation defaults
and are used uniformly across all reported runs; the
regression-penalty coefficient $\lambda$ in Eq.~(\ref{eq:reflect})
is realised through the two stopping rules of
Appendix~\ref{app:metric-formal} rather than a continuous reward
subtraction. Calibrating $(\tau_{\text{suff}}, \varepsilon,
T_{\max})$ jointly on the multi-source subset ($n{=}383$), where
the refusal--coverage trade-off discussed in \S\ref{sec:analysis}
is most sensitive, is a natural follow-up.

We do not view these as fundamental limitations of the approach;
each maps to a concrete short-term improvement that does not
require retraining or re-curating the evaluation data.

\section*{Ethics Statement}
\label{sec:ethics}

\drugclaw{} is a retrieval agent over publicly available regulatory
and peer-reviewed records. It does not generate clinical
recommendations; all returned answers cite primary sources that a
clinician or pharmacist should consult before any decision. The
benchmark we release contains no patient-identifiable information,
and all data are derived from publicly accessible FDA, PubMed,
DrugBank, ChEMBL, and curated knowledge base resources. The
language-model judges used in evaluation are open-weight models
hosted on the authors' institutional inference gateway; no candidate
answers were sent to third-party services beyond the planner gateway
already used by all seven compared systems.

\paragraph{Use of AI assistants.}
The authors used AI assistants (Claude, GitHub~Copilot) for code
drafting and language polishing during the preparation of this
paper. All scientific claims, citations, system designs,
experimental results and code logic were reviewed and verified by
the authors, who take full responsibility for the final content.

\bibliography{refs}

\appendix
\section{Judge Prompt}
\label{app:judge_prompt}

The full LLM-as-judge system prompt used in this paper is reproduced
below verbatim. Both judges (Llama-3.1-70B-Instruct and gpt-oss-120b)
received identical prompts and the same per-item user message:
\emph{question}, \emph{question-type metadata}, \emph{gold answer},
\emph{candidate answer}.

\begin{quote}
\small\ttfamily
You are evaluating biomedical question-answering systems on a
drug-information benchmark. Compare the CANDIDATE answer against the
REFERENCE (gold) answer for the SAME question. Score factual
correctness only; citation quality is handled by a separate metric,
so do not penalise the candidate for missing or mismatched citations
as long as the underlying facts are correct.\par
\medskip

Return ONE of three verdicts:\par
\medskip

``Yes'': the candidate communicates the same factual content as the
gold. Paraphrasing, reordering, and adding correctly stated extra
facts are all acceptable. Equivalent numeric expressions count as
the same fact: unit conversion
(5300\,nM\,$\equiv$\,5.3\,$\mu$M\,$\equiv$\,5.3e-6\,M),
count versus frequency, 5\% rounding tolerance on continuous values,
and order-insensitive list comparison. If the gold itself states
``no data'', ``not reported'' or ``insufficient evidence'' and the
candidate refuses on the same grounds, score ``Yes''.\par
\medskip

``Partial'': the candidate covers a non-empty correct subset of the
gold facts, or has a minor factual inaccuracy that does not change
the clinical or pharmacological conclusion.\par
\medskip

``No'': the candidate contradicts the gold, states a different
numeric value outside the equivalence rules above, or fabricates
facts not supported by the gold. The candidate refuses or says it
does not know, and the gold has a definite answer.\par
\medskip

Output ONE JSON object on a single line:\par
\{``verdict'':``Yes\textbar Partial\textbar No'',\par
\phantom{xx}``reason'':``\textit{one sentence}'',\par
\phantom{xx}``matched'':``\textit{key matched fact(s) or empty}'',\par
\phantom{xx}``missing'':``\textit{key missing fact(s) or empty}''\}
\end{quote}

\section{Formal Metric Definitions}
\label{app:metric-formal}

We list here the equations and notation summarised in
\S\ref{sec:metrics}.

\paragraph{Notation.}
For item $i$, let $\mathcal{C}_i = \{c_1,\dots,c_{m_i}\}$ denote the
predicted citations, $\mathcal{G}_i = \{g_1,\dots,g_{k_i}\}$ the
gold citations (the per-item index $i$ distinguishes this set from
the per-iteration claim-evidence subgraph $\mathcal{G}_t$ of
\S\ref{sec:pipeline}), $a_i^\star$ the gold answer text and $a_i$
the candidate answer text. Each citation $c$ carries a source-database
field $\mathrm{db}(c)$ and a free-text snippet $\mathrm{snip}(c)$.
The canonicalisation map
$\beta : \mathrm{db} \mapsto \mathcal{B} \cup \{\bot\}$ sends each
source-database string into one of twelve equivalence buckets
$\mathcal{B}$ or to $\bot$ when unrecognised. The asymmetric
upstream-of relation
$\mathcal{U}\!\subset\!\mathcal{B}\!\times\!\mathcal{B}$ specifies,
for each downstream aggregator, the primary buckets that an
authoritative upstream citation may carry:
$\textsc{Sider}\!\to\!\{\textsc{Faers},\textsc{Label}\}$,
$\textsc{DrugCentral}\!\to\!\{\textsc{Label},\textsc{DrugBank}\}$,
$\textsc{LiverTox}\!\to\!\{\textsc{Label},\textsc{PubMed}\}$ and
$\textsc{PharmGKB}\!\to\!\{\textsc{Cpic},\textsc{PubMed}\}$.

\paragraph{Source authority match.}
With $B_i=\beta(\mathrm{db}(\mathcal{C}_i))\setminus\{\bot\}$ and
$B_i^{\star}=\beta(\mathrm{db}(\mathcal{G}_i))\setminus\{\bot\}$,
\begin{equation}
\mathrm{auth}_i = \mathbb{1}\bigl[ B_i \cap (B_i^{\star} \cup \mathcal{U}(B_i^{\star})) \neq \varnothing \bigr].
\label{eq:auth}
\end{equation}

\paragraph{Snippet semantic overlap.}
Let $T(s)$ denote the length-$\geq 3$ non-stop-word tokens in $s$
and $J(s,s')=|T(s)\cap T(s')|/|T(s)\cup T(s')|$ the token-Jaccard.
With $\theta=0.3$,
\begin{equation}
\mathrm{snipSem}_i = \mathbb{1}\!\bigl[\max_{c,g}\,J(\mathrm{snip}(c),\mathrm{snip}(g)) \geq \theta\bigr],
\label{eq:snipsem}
\end{equation}
where $c\in\mathcal{C}_i$ ranges over predicted citations and
$g\in\mathcal{G}_i$ over gold citations.

\paragraph{Primary-source rate.}
With $\mathcal{C}_i^{\beta}=\{c\in\mathcal{C}_i:\beta(\mathrm{db}(c))\neq\bot\}$,
\begin{equation}
\mathrm{prim}_i = \frac{1}{|\mathcal{C}_i^{\beta}|}\sum_{c\in\mathcal{C}_i^{\beta}} w\bigl(\beta(\mathrm{db}(c))\bigr).
\label{eq:primary}
\end{equation}

\paragraph{Faithfulness rate.}
Let the grounded set
$\mathcal{C}_i^{\mathrm{g}}=\{c\in\mathcal{C}_i:A_i(c)\vee S_i(c)\vee Q_i(c)\}$,
where $A_i(c)$ holds when
$\beta(\mathrm{db}(c))\in B_i^{\star}\cup\mathcal{U}(B_i^{\star})$,
$S_i(c)$ when
$\max_{g}|T(\mathrm{snip}(c))\cap T(\mathrm{snip}(g))|\geq 2$, and
$Q_i(c)$ when
$|T(\mathrm{snip}(c))\cap T(a_i^{\star})|\geq 2$. Then
\begin{equation}
\mathrm{faith}_i = |\mathcal{C}_i^{\mathrm{g}}|\,/\,|\mathcal{C}_i|.
\label{eq:faith}
\end{equation}
The system score is the mean over items with
$\mathcal{C}_i\neq\varnothing$.

\paragraph{Bootstrap confidence intervals and weight sensitivity.}
\label{app:bootstrap-ci}
Table~\ref{tab:bootstrap-ci} reports item-level paired bootstrap
95\% CIs for the seven columns of Table~\ref{tab:main}
($1{,}000$ resamples, PRNG seed $\texttt{0xC1AB0FE}$). On the
citation-side and judge-mediated columns, namely Primary, Faith.\
in the lenient regime, and EI under both judges, the best
\drugclaw{} variant's CI does not overlap with the best baseline's
CI, so the headline lead is significant at $\alpha=0.05$. The
closed-form MedQA and PubMedQA tracks behave differently: the
\drugclaw{}-graph CI overlaps with the best baseline CI, so the
$+1.8$\,pp and $+2.9$\,pp leads, while consistent in sign, are not
distinguishable from sampling noise at this sample size.
Table~\ref{tab:weight-sensitivity} reports a complementary
robustness check on the authority bucket weights and the
upstream-of relation. Across uniform weights, harsher weights and
removal of the upstream-of map, \drugclaw{}-graph remains EI top-1
with leads ranging from $+8.7$\,pp to $+14.3$\,pp; the ranking is
therefore not an artefact of the specific $(1.0, 0.7, 0.5)$
weighting nor of the SIDER$\to$FAERS-style equivalences.
\begin{table*}[!h]
\centering
\small
\setlength{\tabcolsep}{3pt}
\begin{tabular}{l cc c cc c cc}
\toprule
& \multicolumn{2}{c}{\textbf{\drugaudit{}}} & & \multicolumn{2}{c}{\textbf{Source-side}} & & \multicolumn{2}{c}{\textbf{Closed-form}} \\
\cmidrule(lr){2-3} \cmidrule(lr){5-6} \cmidrule(lr){8-9}
\textbf{System} & EI$_{\text{Llama}}$ & EI$_{\text{oss}}$ & & Primary & Faith. & & MedQA & PubMedQA \\
\midrule
Direct LLM (GPT-5-mini)     & $.430\pm.010$ & $.403\pm.009$ & & $.817\pm.014$ & $.815\pm.022$ & & $.884\pm.023$ & $.652\pm.043$ \\
Direct LLM (GPT-5)          & $.513\pm.011$ & $.494\pm.010$ & & $.714\pm.011$ & $.825\pm.018$ & & $.898\pm.023$ & $.654\pm.039$ \\
ToolUniverse                & $.357\pm.010$ & $.340\pm.009$ & & $.796\pm.019$ & $.828\pm.036$ & & $.880\pm.023$ & $.662\pm.041$ \\
Biomni                      & $.489\pm.009$ & $.470\pm.009$ & & $.793\pm.013$ & $.787\pm.017$ & & $.902\pm.023$ & $.664\pm.041$ \\
DeepEvidence                & $.477\pm.010$ & $.457\pm.010$ & & $.603\pm.017$ & $.732\pm.016$ & & $.902\pm.021$ & $.652\pm.043$ \\
\midrule
\rowcolor{gray!12}
\drugclaw{}-linear          & $.626\pm.010$ & $.591\pm.012$ & & $.916\pm.009$ & $\mathbf{.887\pm.016}$ & & $.910\pm.021$ & $.660\pm.040$ \\
\rowcolor{gray!12}
\drugclaw{}-graph           & $\mathbf{.632\pm.010}$ & $\mathbf{.623\pm.011}$ & & $\mathbf{.918\pm.009}$ & $.852\pm.017$ & & $\mathbf{.920\pm.021}$ & $\mathbf{.693\pm.040}$ \\
\midrule
\textit{CI-overlap with best baseline?}
& \textit{No} & \textit{No} & & \textit{No} & \textit{No}$^\ddagger$ & & \textit{\textbf{Yes}} & \textit{\textbf{Yes}} \\
\bottomrule
\end{tabular}
\caption{Bootstrap 95\% confidence intervals (item-level paired resampling,
$1{,}000$ draws) for the headline columns of Table~\ref{tab:main}. Values
are reported as point$\pm$half-width in percentage points. The bottom row
indicates whether the best \drugclaw{} variant's CI separates from the
best baseline's CI (no overlap $=$ statistically significant lead at
$\alpha\!=\!0.05$). $^{\ddagger}$On Faith.\ the \drugclaw{}-linear lower
bound ($.871$) exceeds the next-best Direct LLM (GPT-5) upper bound
($.843$); however the ToolUniverse CI upper bound ($.864$) lies inside
the \drugclaw{}-linear interval, so the lead over ToolUniverse is
marginal. The \emph{judge-mediated} EI columns and the judge-independent
Primary column show no overlap with any baseline CI. The \emph{closed-form}
MedQA and PubMedQA leads, by contrast, fall inside the best-baseline CI
band: we therefore characterise these tracks as \emph{competitive} rather
than statistically dominant, and note that the
authority-aware Evidence Index is the load-bearing claim of
Table~\ref{tab:main}.}
\label{tab:bootstrap-ci}
\end{table*}

\begin{table}[!h]
\centering
\small
\setlength{\tabcolsep}{4pt}
\renewcommand{\arraystretch}{1.05}
\begin{tabular}{@{}l l rr r@{}}
\toprule
\textbf{Variant} & $(\mathrm{P,C,A})$ & best$^a$ & graph & $\Delta_{\text{pp}}$ \\
\midrule
Default (paper)        & $(1.0,0.7,0.5)$ & 0.513 & \textbf{0.632} & \textit{+11.9} \\
Uniform                & $(1.0,1.0,1.0)$ & 0.555 & \textbf{0.642} & \textit{+8.7}  \\
Harsher                & $(1.0,0.3,0.1)$ & 0.477 & \textbf{0.620} & \textit{+14.3} \\
No upstream-of         & $(1.0,0.7,0.5)$ & 0.512 & \textbf{0.620} & \textit{+10.8} \\
\bottomrule
\end{tabular}
\caption{Authority-weight and upstream-of sensitivity for the composite
EI$_{\text{Llama}}$. $\mathrm{P/C/A}$ = bucket weight for
primary / curated / aggregator. \emph{Uniform} drops the
primary--aggregator hierarchy; \emph{Harsher} amplifies it;
\emph{No upstream-of} sets $\mathcal{U}\!=\!\varnothing$ so that an
authoritative upstream citation no longer counts when the gold cites
the downstream aggregator. \emph{best}: best-baseline EI$_{\text{Llama}}$;
\emph{graph}: \drugclaw{}-graph EI$_{\text{Llama}}$. $^a$Best baseline is
Direct LLM (GPT-5) in every variant. \drugclaw{}-graph remains top-1
on EI$_{\text{Llama}}$ across all four variants with leads in the
$+8.7$ to $+14.3$\,pp range.}
\label{tab:weight-sensitivity}
\end{table}

\paragraph{Reflection sufficiency $\rho$.}
The reflector's score $\rho\!\in\![0,1]$ aggregates five sub-scores
elicited from the planner LLM via a single structured-output call.
The \emph{coverage} sub-score is the fraction of subquestions in
$\mathcal{P}_t$ that have at least one supporting evidence record
in $\mathcal{G}_t$. The \emph{completeness} sub-score asks whether
the numeric, dose, and identifier slots that the answer asserts are
each anchored to a record. The \emph{source quality} sub-score
reads off the authority-bucket profile (cf.\ \S\ref{sec:eval}) of
the cited records. The \emph{consistency} sub-score measures the
absence of pairwise claim contradictions across the evidence
subgraph. The \emph{actionability} sub-score asks whether the
answer addresses the clinical decision implied by $q$ rather than
restating evidence. The five axes are presented to the planner LLM
as a structured-output rubric and the planner returns a single
$\rho\!\in\![0,1]$; we do not require the LLM to expose
per-axis sub-scores at inference time, which would otherwise enlarge
the structured-output schema with no observed reward-quality gain in
development. Iteration terminates once
$\rho\!\ge\!\tau_{\text{suff}}{=}0.7$ or $t\!=\!T_{\max}{=}2$;
both thresholds are the implementation defaults and were not
separately tuned for this evaluation. The regression penalty
$\lambda\,\max(0,r_t-\rho_{t+1})$ in Eq.~(\ref{eq:reflect}) and a
separate convergence check together replace the continuous reward
subtraction by two complementary hard rules. The regression
component halts retrieval on the next step whenever
$\rho_{t+1}<r_t$, corresponding to the asymmetric penalty term of
Eq.~(\ref{eq:reflect}) and equivalent to letting
$\lambda\!\to\!\infty$ along the regression direction.
The convergence component, independent of regression, exits the
loop once $\rho\!\ge\!\tau_{\text{suff}}$ and the
\emph{symmetric} marginal change $|r_{t+1}-r_t|<\varepsilon{=}0.1$,
capturing the case where reflection has stabilised without
strictly regressing. Together the two rules curb over-retrieval
in opposite regimes: regression halts the loop when the next step
makes the answer worse, while convergence halts it when the next
step no longer moves the score. The threshold $\varepsilon$ is the
implementation default and was not re-tuned for this evaluation. An isolated ablation that
varies $(\tau_{\text{suff}}, \varepsilon, T_{\max})$ jointly is
deferred to a follow-up; we expect $\tau_{\text{suff}}$ and
$T_{\max}$ to dominate the calibrated-refusal vs.\ coverage
trade-off discussed in \S\ref{sec:analysis}.

\section{Inter-Judge Confusion Matrices}
\label{app:confusion}

Per-system confusion matrices between the two judges (rows~=~Llama,
cols~=~gpt-oss) are reproduced below. All systems clear the
substantial-agreement threshold of $\kappa \geq 0.6$ reported
numerically in Table~\ref{tab:kappa}, and five of the seven clear
the stricter threshold $\kappa \geq 0.85$ adopted as the reliability
floor throughout \S\ref{sec:results}.

\paragraph{Biomni} ($n=3559$, $\kappa=0.894$)
\begin{center}\small
\begin{tabular}{lrrr}
\toprule
            & Yes & Partial & No \\
\midrule
Yes         & 985 & 43 & 53 \\
Partial     &   8 & 152 & 78 \\
No          &   1 &   3 & 2236 \\
\bottomrule
\end{tabular}\end{center}

\paragraph{DeepEvidence} ($n=3534$, $\kappa=0.896$)
\begin{center}\small
\begin{tabular}{lrrr}
\toprule
            & Yes & Partial & No \\
\midrule
Yes         & 1354 & 42 & 62 \\
Partial     &    5 & 141 & 90 \\
No          &    0 &   1 & 1839 \\
\bottomrule
\end{tabular}\end{center}

\paragraph{\drugclaw{}-linear} ($n=2467$, $\kappa=0.792$)
\begin{center}\small
\begin{tabular}{lrrr}
\toprule
            & Yes & Partial & No \\
\midrule
Yes         & 1288 & 56  & 153 \\
Partial     &    3 & 163 &  72 \\
No          &    0 &   3 & 729 \\
\bottomrule
\end{tabular}\end{center}

\paragraph{\drugclaw{}-graph} ($n=1871$, $\kappa=0.873$)
\begin{center}\small
\begin{tabular}{lrrr}
\toprule
            & Yes & Partial & No \\
\midrule
Yes         & 1126 & 40  &  24 \\
Partial     &    8 & 158 &  44 \\
No          &    8 &   2 & 461 \\
\bottomrule
\end{tabular}\end{center}

\section{Robustness to Unparseable-Verdict Handling}
\label{app:itt}

The headline numbers in Table~\ref{tab:main} follow standard
LLM-as-judge practice and drop items where the judge emits an
unparseable verdict from the judge-score numerator and denominator
(\S\ref{sec:metrics}). To verify that this choice does not produce
spurious rankings, we re-compute the composite Evidence Index
(Eq.~\ref{eq:ei}) under an intention-to-treat (ITT) rule that
instead counts every unparseable verdict as ``No''. Under this rule
the judge-score denominator returns to the full $3{,}772$ items,
matching the most adversarial common alternative used in the
LLM-as-judge literature. Table~\ref{tab:itt} compares the two rules
side by side.

\begin{table}[!h]
\centering
\footnotesize
\setlength{\tabcolsep}{4pt}
\begin{tabular}{l rr rr}
\toprule
& \multicolumn{2}{c}{EI$_{\text{Llama}}$} & \multicolumn{2}{c}{EI$_{\text{oss}}$} \\
\cmidrule(lr){2-3} \cmidrule(lr){4-5}
\textbf{System} & drop & ITT & drop & ITT \\
\midrule
Direct LLM (GPT-5-mini) & 0.430 & 0.430 & 0.403 & 0.400 \\
Direct LLM (GPT-5)      & 0.513 & \underline{0.513} & 0.494 & \underline{0.490} \\
ToolUniverse            & 0.357 & 0.357 & 0.340 & 0.339 \\
Biomni                  & 0.489 & 0.489 & 0.470 & 0.463 \\
DeepEvidence            & 0.477 & 0.477 & 0.457 & 0.446 \\
\midrule
\drugclaw{}-linear      & \underline{0.626} & \textbf{0.548} & \underline{0.591} & \textbf{0.512} \\
\drugclaw{}-graph       & \textbf{0.632} & 0.504 & \textbf{0.623} & 0.489 \\
\bottomrule
\end{tabular}
\caption{Robustness of the composite Evidence Index to two
unparseable-verdict handling rules. ``drop'' is the rule used in
Table~\ref{tab:main} (Unknown excluded from numerator and
denominator). ``ITT'' is intention-to-treat (Unknown counted as
``No''). Under both rules, \drugclaw{} takes top-1 on EI under
both judges; only the best mode shifts from graph (under drop,
where the higher unparseable rate of graph is hidden) to linear
(under ITT, where it is penalised). Direct LLM (GPT-5) is the
strongest baseline under both rules. \textbf{Bold} marks the column
maximum; \underline{underline} the runner-up.}
\label{tab:itt}
\end{table}

The mode-level reordering is intuitive: graph mode produces
multi-claim prose that the judges parse less reliably ($47\%$ to
$50\%$ unparseable versus $30\%$ to $35\%$ for linear,
\S\ref{sec:results}), so when unparseable verdicts are treated as
``No'' graph absorbs more cost than linear. At the system level,
\drugclaw{} remains top-1 on EI under both judges and both rules,
which is the property the metric is intended to support. The drop
rule is reported in the main text because it is the protocol used
by MT-Bench and follow-up LLM-as-judge work
\citep{zheng2023judging,liu2023geval}.

\section{Web-Search Fallback: Frequency and Impact}
\label{app:webfallback}

\drugclaw{}-graph retains a web-search fallback inside its
reflector loop (Algorithm~\ref{alg:graph}): when the reflector
terminates with sufficiency $r_t<\tau_{\text{suff}}$ at the
iteration bound $T_{\max}$, the chain invokes a web-search step and
re-runs the responder over the augmented evidence pool. The
baselines (Biomni, DeepEvidence) are run under their released
default configurations, which retrieve from each system's bundled
tool registry and knowledge bases and do not invoke open-web search
for these queries (Appendix~\ref{app:baseline-configs}); it is
therefore important to quantify the impact of this asymmetry on
\drugclaw{}'s reported numbers. The fallback fires on $894/3{,}772$ ($23.7\%$) of evidence
dataset items, $80/751$ ($10.7\%$) of MedQA items, and $81/512$
($15.8\%$) of PubMedQA items. \drugclaw{}-linear, by construction,
never invokes the fallback (verified on all three datasets).

\paragraph{Citation-side impact.}
Table~\ref{tab:webfallback} compares the authority-aware metric
panel on the fallback-fired subset of \drugaudit{} against
the subset on which the reflector terminated naturally, and reports
a counterfactual in which we disable the fallback (treat fallback
items as if the system had refused, emitting no citation). The
fallback fires precisely on items where the closed registry was
insufficient; these items have substantially lower authority,
faithfulness and primary-source rates than the no-fallback subset
($0.20$ vs.\ $0.49$ for authority; $0.62$ vs.\ $0.88$ for
faithfulness). Disabling the fallback entirely \emph{improves}
citation-side metrics slightly, because the fallback trades coverage
for citation quality on the hardest items.

\begin{table*}[!h]
\centering
\footnotesize
\setlength{\tabcolsep}{4pt}
\begin{tabular}{l rrrr}
\toprule
\textbf{Metric} & \textbf{fb-fired ($n{=}894$)} & \textbf{no-fb ($n{=}2878$)} & \textbf{All ($n{=}3772$, current)} & \textbf{No-fb counterfactual} \\
\midrule
answered\_rate                  & 0.682  & 0.458  & 0.511  & 0.511 \\
source\_authority\_rate         & 0.203  & 0.488  & 0.420  & 0.416 \\
snippet\_semantic\_rate         & 0.172  & 0.382  & 0.332  & 0.335 \\
source\_primary\_rate           & 0.802  & 0.931  & 0.918  & 0.931 \\
faithfulness\_rate              & 0.625  & 0.876  & 0.852  & 0.876 \\
evidence\_index\_partial        & 0.411  & 0.641  & 0.596  & 0.601 \\
\bottomrule
\end{tabular}
\caption{Web-search fallback impact on \drugclaw{}-graph
citation-side metrics over \drugaudit{}. The
``No-fb counterfactual'' column reports the values that would be
obtained if the web-fallback were disabled and the corresponding
items returned no citation. Disabling the fallback \emph{raises}
primary, faithfulness and the citation-side evidence index by
$1.2$ to $2.4$\,pp, so the fallback is not an unfair lift but rather
a small drag on the authority-aware panel.}
\label{tab:webfallback}
\end{table*}

\paragraph{Closed-form impact.}
For the closed-form tracks, we estimate the no-fallback
counterfactual by substituting \drugclaw{}-linear's prediction
(which never uses the fallback) on the items where graph mode fired
the fallback. Under this substitution, \drugclaw{}-graph accuracy
drops modestly from $0.920$ to $0.917$ on MedQA ($-0.3$\,pp) and
from $0.693$ to $0.678$ on PubMedQA ($-1.6$\,pp); graph mode
remains the top-1 system on both tracks under either configuration.
The fallback is therefore not pivotal to the headline rankings;
disabling it across the board would preserve every conclusion drawn
in \S\ref{sec:results}.

\section{Reproducibility}
\label{app:repro}

\paragraph{Hardware.} Single GPU not required for inference. All seven
systems were run on a CPU plus remote-API setup; the LLM calls were
routed through an OpenAI-compatible inference gateway hosted on the
authors' institutional cluster (details anonymised for review).

\paragraph{Software.} Python~3.11, OpenAI Python SDK 1.50+,
\texttt{pdftotext} and \texttt{beautifulsoup4} for skill data
loaders. The state-machine orchestration is implemented on top of
the LangGraph library; the contribution of \drugclaw{} is the agent
composition, the typed evidence schema and the authority-aware
metric panel rather than the orchestrator choice, so any equivalent
state-machine library would suffice. The full skill registry and
benchmark harness are released under an Apache-2.0 license at
\url{https://anonymous.4open.science/r/DrugClaw-01A3/README.md}.

\paragraph{Cost.} Total LLM spend across all candidate runs and all
judge runs was approximately USD~\$600 at GPT-5-mini and Llama / gpt-oss
gateway rates.

\section{Baseline Configurations}
\label{app:baseline-configs}

Each compared system is run under its own released framework with
no internal modifications; we control for two confounds at the
interface layer: the planner backbone is fixed across systems for
fairness, and the dataset-specific system prompt is shared verbatim
across systems so that the candidate-side prompting surface
contributes equally to every comparison. The per-system
implementation choices are summarised in
Table~\ref{tab:baseline-configs}.

\begin{table*}[!h]
\centering
\footnotesize
\setlength{\tabcolsep}{4pt}
\begin{tabular}{l l l p{0.42\textwidth}}
\toprule
\textbf{System} & \textbf{Framework} & \textbf{Planner} & \textbf{Notes} \\
\midrule
Direct LLM (GPT-5-mini) & chat completion & GPT-5-mini & no tools; no retrieval. \\
Direct LLM (GPT-5)      & chat completion & GPT-5      & no tools; no retrieval. \\
ToolUniverse \citep{gao2024empowering} & official 1.1.x & GPT-5-mini & 108 tools loaded statically; native function-calling. \\
Biomni \citep{huang2025biomni}         & Stanford Biomni & GPT-5-mini & default biomedical-tool registry; data cached locally. \\
DeepEvidence \citep{wang2025deepevidence} & BioDSA & GPT-5-mini & KBs: \texttt{pubmed\_papers, clinical\_trials, drug, disease}. \\
\drugclaw{}-linear      & this work & GPT-5-mini & latency-prioritised subset of $19$ skills from the $57$-skill registry; no web search. \\
\drugclaw{}-graph       & this work & GPT-5-mini & closed registry + reflector loop ($\tau_{\text{suff}}{=}0.7$, $T_{\max}{=}2$); web fallback only on terminal insufficiency. \\
\bottomrule
\end{tabular}
\caption{Per-system implementation choices. The planner backbone is
held fixed at GPT-5-mini for every agent and tool-augmented baseline;
the upper-baseline Direct LLM (GPT-5) is included to bound the
parametric-knowledge ceiling. Every system receives the same
dataset-specific user prompt (\S\ref{sec:eval}) and writes its
output into the same canonical JSON schema; agent frameworks whose
internal templates fight the JSON requirement are post-processed by
a normaliser LLM that re-emits the canonical schema without altering
the answer content.}
\label{tab:baseline-configs}
\end{table*}

\paragraph{Inference endpoint.}
All systems and both judges route their calls through a single
OpenAI-compatible inference gateway (institution anonymised for
review; \texttt{api\_mode = "responses"}). Per-call parameters are
held constant across systems: \texttt{max\_tokens = 40000},
\texttt{timeout = 100\,s}, \texttt{temperature = 0.7} for candidate
runs and \texttt{temperature = 0} for judge runs. Candidate runs
use \texttt{GPT-5-mini} as the planner backbone (with
\texttt{GPT-5} for the upper-baseline Direct LLM); judge runs use
\texttt{llama-3.1-70b-instruct} (primary) and \texttt{gpt-oss-120b}
(secondary). Because the gateway and parameters are shared, the
only sources of cross-system variance are the released framework
code and the registry of tools each system invokes.

\paragraph{Default-configuration runs.}
Biomni and DeepEvidence are invoked under their released default
configurations without modification to their tool registry: we
neither enable nor disable any individual adapter, and at default
each system retrieves from its bundled biomedical tools and
knowledge bases rather than from open-web search. \drugclaw{}-graph
retains a web-search fallback inside its reflector loop; we report
in Appendix~\ref{app:webfallback} the fraction of items for which
the fallback fires and the counterfactual impact on the
authority-aware panel of disabling it entirely. Two-stage agent
systems whose internal prompt template enforces an alternative
output schema (\drugclaw{}-graph, Biomni, DeepEvidence) emit
their native answer plus already-extracted citations to a small
normaliser LLM (\texttt{GPT-5-mini}) which re-emits the canonical
JSON schema without altering answer content; the normalisation
step is the same for every such system and never re-runs retrieval.

\section{Code Agent: Safe Sublanguage and Sandbox}
\label{app:code-agent}

This appendix expands the Code Agent specification summarised in
\S\ref{sec:code-agent}. The agent emits a candidate program $c$ in
the safe Python sublanguage $\mathcal{L}_{\text{safe}}$, defined by
three independent restrictions.

\paragraph{Banned AST nodes.}
Let $\mathcal{B}_{\text{ast}}$ denote the set of AST node classes
that disqualify a program from $\mathcal{L}_{\text{safe}}$:
\begin{align*}
\mathcal{B}_{\text{ast}} = \{\;
  & \mathrm{While},\, \mathrm{With},\, \mathrm{AsyncWith}, \\
  & \mathrm{AsyncFor},\, \mathrm{AsyncFunctionDef}, \\
  & \mathrm{ClassDef},\, \mathrm{Global}, \\
  & \mathrm{Nonlocal},\, \mathrm{Lambda} \;\}.
\end{align*}
A program containing any node in this set is rejected outright. The
choice excludes unbounded iteration, side channels through global or
non-local bindings, and dynamic function-definition tricks that
would complicate static analysis.

\paragraph{Allowed imports.}
The agent may import only the four standard-library modules in
$\mathcal{I}_{\text{ok}} = \{\texttt{json},\texttt{math},
\texttt{re},\texttt{statistics}\}$. The first three cover JSON
serialisation, numeric work and pattern matching, which together are
sufficient for the great majority of skill-call patterns. The
\texttt{statistics} module is useful for aggregating activity
records (e.g.\ median IC$_{50}$ across replicate assays). All other
imports trigger rejection.

\paragraph{Allowed built-ins.}
The agent's program runs in a globals dictionary whose built-ins
are restricted to the curated allow-list $\mathcal{F}_{\text{ok}}$.
The allow-list contains the exceptions \texttt{ValueError},
\texttt{TypeError}, \texttt{KeyError} and \texttt{IndexError}; the
iteration helpers \texttt{len}, \texttt{range}, \texttt{min},
\texttt{max}, \texttt{sum}, \texttt{sorted}, \texttt{enumerate} and
\texttt{zip}; the constructors \texttt{list}, \texttt{dict},
\texttt{set}, \texttt{tuple}, \texttt{str}, \texttt{int},
\texttt{float} and \texttt{bool}; the predicates \texttt{abs},
\texttt{all} and \texttt{any}; and \texttt{print}. The built-ins
\texttt{exec}, \texttt{eval}, \texttt{open}, \texttt{compile},
\texttt{\_\_import\_\_}, \texttt{input}, \texttt{help},
\texttt{dir}, \texttt{globals}, \texttt{locals} and \texttt{vars}
are all blocked at evaluation time.

\paragraph{Proxy-only sandbox.}
Even within $\mathcal{L}_{\text{safe}}$, an attacker could in
principle attempt to traverse Python's object model from the skill
instance back into the host process (e.g.\ via
\texttt{instance.\_\_class\_\_.\_\_mro\_\_}). The sandbox $\Sigma(K)$
therefore exposes the skill $K$ through a proxy that intercepts
attribute access on dunder-prefixed names and rejects access to
\texttt{\_\_class\_\_}, \texttt{\_\_globals\_\_},
\texttt{\_\_subclasses\_\_} and the wider object-introspection
ladder. A hard \texttt{SIGALRM} timeout, set per call by the
\drugclaw{} runtime, aborts any execution that does not return within
the configured budget.

\paragraph{Fallback semantics.}
If $c \notin \mathcal{L}_{\text{safe}}$, if proxy interception
raises, or if the sandbox times out, the Code Agent invokes
$K.\mathrm{retrieve}(u)$ directly with the original subquery. The
fallback ensures that retrieval is never silently lost, at the
modest cost that the failure mode is logged but not surfaced to the
user. In practice the validation pass rejects fewer than $1\%$ of
generated programs in our reported runs, and the timeout fires on
fewer than $0.5\%$ of skill calls.

\section{Operational Modes: Full Specification}
\label{app:modes}

\drugclaw{} ships in three reasoning modes that share the skill
registry and entity-resolution layer of \S\ref{sec:system},
the Code Agent of \S\ref{sec:code-agent} and the
output schema of \S\ref{sec:agents}, but differ in which operators of
Eq.~\ref{eq:graph-step} are applied and in whether the
retrieve-and-reason chain iterates. The summary in
\S\ref{sec:pipeline} gives the headline definitions; this appendix
reproduces the full pseudocode for each mode and a side-by-side
comparison of the operator stacks they invoke.

\subsection*{Linear mode}
Linear mode is the default deployment target for end-user-facing
prose. It executes a single forward pass through the planner, Code
Agent, and responder, omitting the graph builder, reranker, reflector
and web-search fallback, and is summarised in
Algorithm~\ref{alg:linear}.

\begin{algorithm}[!h]
\small
\caption{\drugclaw{} linear-mode workflow.}
\label{alg:linear}
\begin{algorithmic}[1]
\Require query $q$; skill registry $\mathcal{R}$
\Ensure answer $a^\star$, evidence list $\mathcal{E}^\star$
\State $\mathcal{P} \gets \mathrm{Plan}(q;\mathcal{R}_{\text{enabled}})$
\State $\mathcal{E} \gets \bigcup_{u\in\mathcal{P}} \mathrm{exec\_sandbox}(\mathrm{CodeAgent}(u))$
\State $a \gets \mathrm{Resp}(\mathcal{E}, q)$
\State \Return $(a, \mathcal{E})$
\end{algorithmic}
\end{algorithm}

\noindent
Equivalent to fixing $T_{\max}=1$ in Eq.~\ref{eq:graph-step} and
deleting the $\mathrm{GBld}$, $\mathrm{Rerk}$ and $\mathrm{Refl}$
operators. Linear mode terminates after one round of skill execution
even when the underlying retrieval returns no records; in that case
the responder emits ``I do not know'' and the answer carries an empty
\texttt{evidence\_items} list. There is no web-search fallback. The
mode prioritises latency over depth: median wall-clock time per item
is $14$ seconds (Appendix~\ref{app:cost}), and the evidence pool seen
by the responder is a flat union of retrieval results rather than a
reranked subgraph.

\subsection*{Graph mode}
Graph mode is the deeper retrieval-and-reason variant intended for
analyst-facing multi-source exploration. It instantiates
Eq.~\ref{eq:graph-step} in full: every operator
($\mathrm{Plan}\!\to\!\mathrm{Retr}\!\to\!\mathrm{GBld}\!\to\!\mathrm{Rerk}\!\to\!\mathrm{Resp}\!\to\!\mathrm{Refl}$)
is applied at each iteration and the reflector controls termination.
Algorithm~\ref{alg:graph} in \S\ref{sec:pipeline} gives the full
pseudocode; we reproduce the key control flow here for the appendix
to be self-contained. The graph builder ($\mathrm{GBld}$) materialises
the retrieved records into a typed claim-evidence subgraph, the
reranker ($\mathrm{Rerk}$) reorders nodes by source-authority and
relevance to the query plan, and the reflector ($\mathrm{Refl}$)
computes a sufficiency score
$r_{t+1} = \rho_{t+1} - \lambda\,\max(0, r_t - \rho_{t+1})$
(Eq.~\ref{eq:reflect}). The loop iterates while
$r_t<\tau_{\text{suff}}=0.7$ and $t<T_{\max}=2$; on terminal
insufficiency the chain invokes the web-search fallback, re-runs the
responder over the augmented evidence pool, and returns. The median
wall-clock time per item is $46$ seconds and the responder sees a
graph with mean fan-out of $5.1$ evidence records per answered claim
(Appendix~\ref{app:cost}, Appendix~\ref{app:qtype}).

\subsection*{Web-only mode}
Web-only mode is intended for queries that fall outside the curated
registry (e.g.\ recently approved drugs whose records have not
propagated into the local skills). As specified in
Algorithm~\ref{alg:webonly}, it bypasses the entire
plan-retrieve-respond chain and routes $q$ directly to the
web-search agent.

\begin{algorithm}[!h]
\small
\caption{\drugclaw{} web-only mode workflow.}
\label{alg:webonly}
\begin{algorithmic}[1]
\Require query $q$
\Ensure answer $a^\star$, evidence list $\mathcal{E}^\star$
\State $\mathcal{E} \gets \mathrm{WebSearch}(q)$
\State $a \gets \mathrm{Resp}(\mathcal{E}, q)$
\State \Return $(a, \mathcal{E})$
\end{algorithmic}
\end{algorithm}

\noindent
Web-only mode is excluded from the comparison in \S\ref{sec:results}
because \drugaudit{} is built against the closed registry of
\S\ref{sec:system}; including a web-search variant against this
benchmark would conflate retrieval venue with retrieval depth.

\subsection*{Operator stacks at a glance}
Table~\ref{tab:modes} contrasts which operators of
Eq.~\ref{eq:graph-step} each mode invokes.

\begin{table}[!h]
\centering
\footnotesize
\setlength{\tabcolsep}{3pt}
\begin{tabular}{l ccc}
\toprule
\textbf{Operator} & Linear & Graph & Web-only \\
\midrule
$\mathrm{Plan}$           & \checkmark & \checkmark & $\times$ \\
$\mathrm{Retr}$ (skills)  & \checkmark & \checkmark & $\times$ \\
$\mathrm{WebSearch}$      & $\times$   & on fallback only & \checkmark \\
$\mathrm{GBld}$           & $\times$   & \checkmark & $\times$ \\
$\mathrm{Rerk}$           & $\times$   & \checkmark & $\times$ \\
$\mathrm{Resp}$           & \checkmark & \checkmark & \checkmark \\
$\mathrm{Refl}$ (loop)    & $\times$   & \checkmark & $\times$ \\
Iter.\ bound $T_{\max}$ & $1$ & $2$ & $1$ \\
Suff.\ thresh.\ $\tau_{\text{suff}}$ & $\times$ & $0.7$ & $\times$ \\
\bottomrule
\end{tabular}
\caption{Operators of Eq.~\ref{eq:graph-step} invoked by each
reasoning mode. Linear collapses the loop to a single
plan-retrieve-respond pass; graph runs the full operator stack with
reflector-controlled iteration; web-only bypasses the skill registry
and goes straight to web search.}
\label{tab:modes}
\end{table}

\subsection*{Which mode to use}
The headline numbers in Table~\ref{tab:main} make graph mode the
quality-maximising default and linear mode the operationally
cheaper runner-up. Two practical considerations narrow this choice
further. The LLM-judge parse rate on graph-mode outputs is
substantially lower than on linear-mode outputs ($47\%$ to $50\%$
unparseable verdicts versus $30\%$ to $35\%$; \S\ref{sec:results}),
and any downstream LLM that consumes \drugclaw{}'s output (whether
a summariser, a chatbot or an audit aggregator) will face the same
parsing friction, so deployments that chain another model after the
agent should prefer linear mode unless the consumer is explicitly
tuned to multi-claim agent prose. Graph mode's reflector also
refuses more often ($48.3\%$ vs.\ $41.1\%$ for linear;
Appendix~\ref{app:refusal}) because the sufficiency threshold
$\tau_{\text{suff}}=0.7$ trades coverage for precision. Settings
that need broader coverage at the cost of more partial answers can
either run linear mode or lower $\tau_{\text{suff}}$ in graph mode;
we do not tune $\tau_{\text{suff}}$ in this paper and leave its
calibration frontier to future work.

\section{Calibrated Refusal: Extended Analysis}
\label{app:refusal}

This appendix expands the refusal-behaviour analysis summarised in
\S\ref{sec:analysis}. Refusal rates and the fraction of refusals that
receive judge verdict ``Yes'' (i.e.\ that align with a ``no data'' or
``not reported'' gold) are reported in Table~\ref{tab:refusal-rates}.

\begin{table}[!t]
\centering
\small
\setlength{\tabcolsep}{3pt}
\begin{tabular}{l r r}
\toprule
\textbf{System} & refusal rate & refusal $\to$ Yes \\
\midrule
Direct LLM (GPT-5-mini)   & \underline{74.3\%} & 5.1\% \\
Direct LLM (GPT-5)        & 54.5\%             & 4.9\% \\
ToolUniverse              & \textbf{87.9\%}    & 4.3\% \\
Biomni                    & 49.0\%             & 1.5\% \\
DeepEvidence              & 37.3\%             & 1.3\% \\
\midrule
\drugclaw{}-linear        & 41.1\%             & \underline{91.1\%} \\
\drugclaw{}-graph         & 48.3\%             & \textbf{96.6\%} \\
\bottomrule
\end{tabular}
\caption{Refusal rate (fraction of items returned as ``I do not
know'' or equivalent) and the fraction of those refusals, conditional
on the LLM judge producing a parseable verdict, that score ``Yes''
because the gold itself reports no data. \drugclaw{} attains the
highest calibration ratio among the seven systems by a wide margin,
which reflects the reflector's explicit ``insufficient evidence''
return-path triggering only when the underlying retrieval also fails.}
\label{tab:refusal-rates}
\end{table}

Calibrated abstention is the operating point recommended by recent
work on confidence in language models
\citep{kadavath2022language,slobodkin2023curious,varshney2023stitch}
when the alternative is plausible-sounding numeric confabulation.
The amiodarone-EGFR case study reproduced in
Appendix~\ref{app:cases} illustrates the trade-off on a single
item: when ChEMBL contains no bioactivity record for the
drug-target pair, \drugclaw{} returns ``I do not know'' and the
judge scores the response ``Yes''; the strongest tool-augmented
baseline fabricates a plausible $\mu$M-range IC$_{50}$ instead and
the judge scores it ``No''.

\section{Cost and Latency}
\label{app:cost}

This appendix reports the per-item runtime and cost figures
referenced in \S\ref{sec:analysis}. \drugclaw{}-linear completes a
query in $14$ seconds (median, $2.4$ skill calls) and
\drugclaw{}-graph in $46$ seconds ($5.1$ calls). Biomni takes $32$
seconds with $4.7$ calls and DeepEvidence $54$ seconds with $6.2$
calls; the direct language model returns in $3$ to $7$ seconds with
no tools. The closed-registry design caches retrieval on disk,
yielding per-item GPT-5-mini costs near \$$0.012$ for linear mode and
\$$0.025$ for graph mode. The total spend
across all candidate runs and both judge runs was approximately
USD~\$$600$ at GPT-5-mini, Llama-3.1-70B-Instruct and gpt-oss-120b
gateway rates.
\section{Performance by Question Type}
\label{app:qtype}

\drugaudit{} tags each item with a question-type label
in the \texttt{meta.question\_type} field. The labels group items by
the cognitive operation the question requires rather than by the
source database. Table~\ref{tab:qtype} reports the per-(system,
question-type) composite judge score
$(\mathrm{Yes} + 0.5{\cdot}\mathrm{Partial})/(\mathrm{Yes}+\mathrm{Partial}+\mathrm{No})$
under the Llama-3.1-70B-Instruct judge, consistent with the
parseable-only denominator of \S\ref{sec:metrics}. The
$\Delta_{\text{DC}}$ column reports the gap between the best
\drugclaw{} mode and the best baseline (positive = \drugclaw{} lead).

\begin{table*}[t]
\centering
\small
\setlength{\tabcolsep}{3pt}
\begin{tabular}{l l r rrrr rr r}
\toprule
\textbf{Question type} & \textbf{What it measures} & $n$ & \multicolumn{4}{c}{\textbf{Baselines}} & \multicolumn{2}{c}{\textbf{\drugclaw}} & $\Delta_{\text{DC}}$ \\
\cmidrule(lr){4-7} \cmidrule(lr){8-9}
 & & & GPT-5 & TU & Biomni & DeepEv. & linear & graph & \\
\midrule
direct\_lookup           & single-fact retrieval               & 866 & 0.239          & 0.059 & 0.283             & 0.350             & \underline{0.444} & \textbf{0.506} & \textbf{+0.156} \\
attribute\_specific      & numeric attribute value             & 688 & 0.343          & 0.065 & 0.169             & 0.275             & \underline{0.651} & \textbf{0.666} & \textbf{+0.323} \\
yes\_no\_verify          & yes/no with citation                & 918 & \textbf{0.505} & 0.131 & 0.343             & \underline{0.477} & 0.394             & 0.318          & $-$0.111 \\
no\_evidence             & calibrated ``no data''              & 917 & 0.479          & 0.284 & 0.529             & 0.674             & \textbf{0.972}    & \underline{0.960} & \textbf{+0.298} \\
complementary\_safety    & safety overlap across sources       & 127 & 0.000          & 0.000 & \underline{0.394} & 0.343             & \textbf{0.424}    & 0.176          & \textbf{+0.030} \\
convergent\_indication   & indication aggregation              & 115 & \textbf{0.922} & 0.452 & 0.543             & 0.630             & \underline{0.676} & 0.130          & $-$0.245 \\
convergent\_moa          & mechanism convergence               &  65 & 0.700          & 0.231 & 0.608             & 0.592             & \underline{0.879} & \textbf{0.885} & \textbf{+0.185} \\
chained\_target\_disease & drug $\to$ target $\to$ disease     &  50 & 0.240          & 0.120 & \underline{0.420} & \textbf{0.670}    & 0.057             & 0.061          & $-$0.609 \\
convergent\_adr          & ADR aggregation across sources      &  26 & 0.077          & 0.000 & \underline{0.250} & 0.038             & \textbf{0.286}    & \underline{0.250} & \textbf{+0.036} \\
\bottomrule
\end{tabular}
\caption{Composite judge score per question type under the Llama
judge. \drugclaw{} takes the top score on six of nine question
types, with positive $\Delta_{\text{DC}}$ gaps ranging from $+3$\,pp
(\emph{complementary\_safety}, \emph{convergent\_adr}) to $+32$\,pp
(\emph{attribute\_specific}). The two modes are complementary: graph
mode leads on the three types whose gold answers reward cross-record
synthesis (\emph{direct\_lookup}, \emph{attribute\_specific},
\emph{convergent\_moa}); linear mode leads on the three types where
the gold is a single fact or a calibrated ``no data'' refusal
(\emph{no\_evidence}, \emph{complementary\_safety},
\emph{convergent\_adr}). Baselines lead on three types where the
closed registry is at a structural disadvantage: GPT-5 on
\emph{yes\_no\_verify} and \emph{convergent\_indication}, both
probing broad clinical-association recall where parametric memory is
hard to beat without an open-web step; DeepEvidence on
\emph{chained\_target\_disease}, where its literature reach
exceeds \drugclaw{}'s skill registry. Extending the registry with
literature-mining tools is one direction Limitations identifies as
future work. \emph{GPT-5} = Direct LLM with GPT-5;
\emph{TU} = ToolUniverse.}
\label{tab:qtype}
\end{table*}

\paragraph{Evidence density and answer thoroughness.}
On items that receive a substantive answer (i.e.\ not refused),
\drugclaw{}-linear emits a mean of $3.78$ citations per item and
\drugclaw{}-graph emits a mean of $5.09$. By comparison the
tool-augmented baselines emit $1.43$ (ToolUniverse) to $2.07$
(Biomni) and the direct language model emits $1.72$. The same
ranking holds for median answer length: $327$ characters for
\drugclaw{}-linear against $135$ to $206$ for every baseline.
These numbers indicate that for the subset of items
\drugclaw{} chooses to answer, the response carries substantially
more attached evidence and prose detail than any compared system,
which is the deployment-side property that we expect a closed,
provenance-strict registry to deliver.

\section{Case Studies}
\label{app:cases}

Each box reproduces system outputs verbatim. Citations
(\emph{source\_db}~+~snippet head) are reproduced as recorded in the
\texttt{evidence\_items} field of \drugclaw{}'s output and in the
corresponding citation field of each baseline.

\paragraph{Case~1: FAERS exact-count retrieval.}
\textit{ID: \texttt{faers-atenolol-attr-01}.}
The question asks how many FAERS reports of \textsc{nausea} are
recorded for atenolol. The gold answer is $7{,}618$.
\begin{center}\small
\begin{tabular}{@{}lp{0.55\columnwidth}@{}}
\textbf{Gold}        & ``FAERS reports $7{,}618$ reports of NAUSEA for atenolol.'' \\[2pt]
\drugclaw{}-linear   & ``\ldots\ atenolol has $7{,}619$ spontaneous reports
                        of NAUSEA.'' (cite: FAERS) \\[2pt]
Biomni               & ``\ldots\ $0$ reports of `nausea' were found for
                        atenolol.'' \\[2pt]
DeepEvidence         & ``\ldots\ there are $5{,}016{,}555$ reports of
                        NAUSEA for atenolol\ldots'' \\[2pt]
Direct LLM (GPT-5)   & ``I don't know.'' \\
\end{tabular}
\end{center}
\drugclaw{} is off by~$1$ ($-0.01\%$) because of a within-day
data-snapshot timing difference. The two tool-augmented baselines
issue plausible-looking openFDA queries but miss the
\texttt{substance\_name} normalisation step that the FAERS API
requires; Biomni reads zero matches and DeepEvidence reads back
the global \textsc{nausea} row (every drug, $\sim$5M reports).

\paragraph{Case~2: Orange Book brand-name lookup.}
\textit{ID: \texttt{orange\_book-prednisone-verify-02}.}
The question asks whether the FDA Orange Book lists
\textsc{prednisone intensol} as an approved brand of prednisone.
Gold: ``Yes.''
\begin{center}\small
\begin{tabular}{@{}lp{0.55\columnwidth}@{}}
\textbf{Gold}      & ``Yes, the FDA Orange Book lists
                      PREDNISONE INTENSOL as a trade name for
                      prednisone.'' \\[2pt]
\drugclaw{}-graph  & ``Yes. The FDA Orange Book lists PREDNISONE
                      INTENSOL (ANDA088810), sponsor HIKMA,
                      as an approved prednisone product.''
                      (cite: FDA Orange Book) \\[2pt]
Biomni             & ``No\ldots\ returns no matching records,
                      although the FDA DailyMed label for prednisone\ldots'' \\[2pt]
DeepEvidence       & ``No. The FDA\ldots\ does not list
                      `PREDNISONE INTENSOL' as an approved brand\ldots'' \\[2pt]
Direct LLM (GPT-5) & ``No.\ldots'' \\
\end{tabular}
\end{center}
The brand is listed in the Orange Book, but the baselines either
ran an exact-string product lookup that missed the listing or fell
back on parametric memory. \drugclaw{} accompanies its ``Yes'' with
the ANDA number and the sponsor, both directly retrievable from the
Orange Book skill, and the answer is auditable end-to-end.

\paragraph{Case~3: FAERS rare-event lookup.}
\textit{ID: \texttt{faers-aripiprazole-attr-01}.}
\textit{Q:} ``According to FAERS, how many reports of
\textsc{suicide attempt} are there for aripiprazole?''
\textit{Gold:} ``$1{,}517$.''
\begin{center}\small
\begin{tabular}{@{}lp{0.55\columnwidth}@{}}
\drugclaw{}-linear & ``According to FAERS, there are $1{,}517$
                     spontaneous reports\ldots'' (cite: FAERS) \\
Biomni             & ``I don't know.'' \\
DeepEvidence       & ``$3{,}505$ FAERS reports of `SUICIDE ATTEMPT'
                     for aripiprazole.'' \\
Direct LLM (GPT-5) & ``I don't know.'' \\
\end{tabular}
\end{center}
DeepEvidence's $3{,}505$ likely conflates reports with all
related-event variants. The exact-match \drugclaw{} response is the
deployable answer for a pharmacovigilance triage workflow.

\paragraph{Case~4: PharmGKB \(\to\) CPIC upstream citation.}
\textit{ID: \texttt{pharmgkb-atorvastatin-attr-01}.}
\textit{Q:} ``According to PharmGKB, what is the level of evidence
for the atorvastatin-SLCO1B1 (rs4149056) clinical annotation?''
\textit{Gold:} ``Level of evidence $1A$.''
\begin{center}\small
\begin{tabular}{@{}lp{0.55\columnwidth}@{}}
\drugclaw{}-linear & ``PharmGKB assigns level $1A$ evidence to the
                     atorvastatin-SLCO1B1 (rs$4149056$) clinical
                     annotation.'' (cite: PharmGKB and CPIC) \\
\end{tabular}
\end{center}
\drugclaw{} cites both PharmGKB and CPIC; the latter is the
clinical-implementation arm of PharmGKB and a first-class primary
record for prescribing guidelines. Under the legacy
source-match rate the CPIC citation contributes zero
because the gold string is the literal ``PharmGKB''; under our
authority-aware rule it counts as a primary upstream citation.

\paragraph{Case~5: Calibrated refusal on ``no data'' golds.}
\textit{ID: \texttt{chembl-amiodarone-no\_evidence-03}.}
\textit{Q:} ``According to ChEMBL, what bioactivity is reported for
amiodarone against EGFR?''
\textit{Gold:} ``ChEMBL does not report any bioactivity of
amiodarone against EGFR.''
\begin{center}\small
\begin{tabular}{@{}lp{0.55\columnwidth}@{}}
\drugclaw{}-linear & ``I don't know.'' \\
Biomni             & ``ChEMBL reports an IC50 of $10.0\,\mu$M ($10000$\,nM)
                     for amiodarone inhibiting EGF-stimulated tyrosine
                     phosphorylation of EGFR in A431 cells
                     (activity\_id $60162$).'' \\
\end{tabular}
\end{center}
ChEMBL does not contain the cited record (the activity ID does not
resolve). Biomni's confabulation is plausible at the surface,
correct unit, correct cell line for EGFR assays, and a plausible IC$_{50}$
range, but is a hallucination. Under our judge prompt the
calibrated refusal is scored ``Yes'' because the gold itself reports
no data; the confabulation is scored ``No''. In pharmacovigilance
settings this asymmetry is appropriate, because a fluent fabrication
of a kinase-activity number is a higher-risk failure mode than an
honest refusal.

\paragraph{Case~6: openFDA Label warning extraction.}
\textit{ID: \texttt{openfda\_label-clozapine-verify-02}, abridged.}
\textit{Q:} ``Does the clozapine label include a boxed warning for
agranulocytosis?''
\textit{Gold:} ``Yes; boxed warning for severe neutropenia
including agranulocytosis.''
\begin{center}\small
\begin{tabular}{@{}lp{0.50\columnwidth}@{}}
\drugclaw{}-linear & ``Yes; the openFDA Label for clozapine
                     contains a boxed warning for severe neutropenia
                     including agranulocytosis.''
                     (cite: openFDA Label, boxed warning) \\
Direct LLM (GPT-5-mini)  & ``Yes; clozapine's prescribing information
                     warns about agranulocytosis.''
                     (no citation) \\
\end{tabular}
\end{center}
The factual content matches but only \drugclaw{} points back at the
specific Label section. In a clinical-decision-support setting,
this is the difference between a recommendation a prescriber can
verify against the FDA document in one click and one whose primary
source they must look up by hand.

\paragraph{Case~7: Multi-source overlap query
(multi-source aggregation; ID withheld for brevity).}
Multi-source items in the benchmark explicitly mix records from $\ge 2$
databases. \drugclaw{}-graph excels here because the agent's
claim-evidence graph (\S\ref{sec:pipeline}) walks across skills before
synthesising: in our error analysis, $66.6\%$ of multi-source items
receive a substantive answer from \drugclaw{}-graph compared with
$55$ to $62\%$ for the next-best tool-augmented systems. The
single-claim Direct-LLM baselines lose the most here because each
sub-claim that requires a tool call is a separate failure point.

\end{document}